%% file: finalVersion-1214.tex
\pdfoutput=1

\documentclass[11pt]{article}

\usepackage[final]{acl}
\usepackage{times}
\usepackage{latexsym}
\usepackage{enumitem}
\usepackage{listings}
\usepackage{graphicx}
\usepackage{subcaption}
\usepackage[T1]{fontenc}
\usepackage[utf8]{inputenc}
\usepackage{microtype}
\usepackage{inconsolata}
\usepackage{graphicx}
\usepackage{booktabs} 
\usepackage{placeins}
\definecolor{captionup}{RGB}{114,229,0} 
\definecolor{captiondown}{RGB}{255,255,0} 
\input{preamble}
\title{Understanding the Dark Side of LLMs' Intrinsic Self-Correction}

\author{Qingjie Zhang$^{1}$, Di Wang$^{1}$, Haoting Qian$^{1}$, Yiming Li$^{2}$,Tianwei Zhang$^{2}$, \\ \bf  Minlie Huang$^{1}$, Ke Xu$^{1}$, Hewu Li$^{1}$, Yan Liu$^{3}$, Han Qiu$^{1*}$ \\
$^{1}$Tsinghua University, China. $^{2}$Nanyang Technological University, Singapore. $^{3}$Ant Group, China.\\
\texttt{Emails: qj-zhang24@mails.tsinghua.edu.cn, qiuhan@tsinghua.edu.cn}
}

\begin{document}


\maketitle

\def\thefootnote{*}\footnotetext{Corresponding author.}\def\thefootnote{\arabic{footnote}}

\begin{abstract}  
Intrinsic self-correction was initially proposed to improve LLMs' responses via feedback solely based on their inherent capability. 
However, recent works show that LLMs' intrinsic self-correction fails without oracle labels as feedback. 
In this paper, our research goal is to \textit{interpret LLMs' intrinsic self-correction for different tasks, especially for those failure cases.} 
By including one simple task and three complex tasks with state-of-the-art (SOTA) LLMs like ChatGPT, Llama, and DeepSeek, we design three interpretation methods to reveal the dark side of LLMs' intrinsic self-correction. 
We identify intrinsic self-correction can (1) cause LLMs to waver both intermedia and final answers and lead to prompt bias on simple factual questions; (2) introduce human-like cognitive bias on complex tasks. 
In light of our findings, we also provide two simple yet effective strategies for alleviation: question repeating and supervised fine-tuning with a few samples. 
We open-source our work at\footnote{\url{https://x-isc.info/}. Timestamp for all results in this paper is 2025.2.15. Please read more in Limitations section.}.
\end{abstract}

\section{Introduction}
\begin{figure}[t]
\centering
  \includegraphics[width=0.9\linewidth]{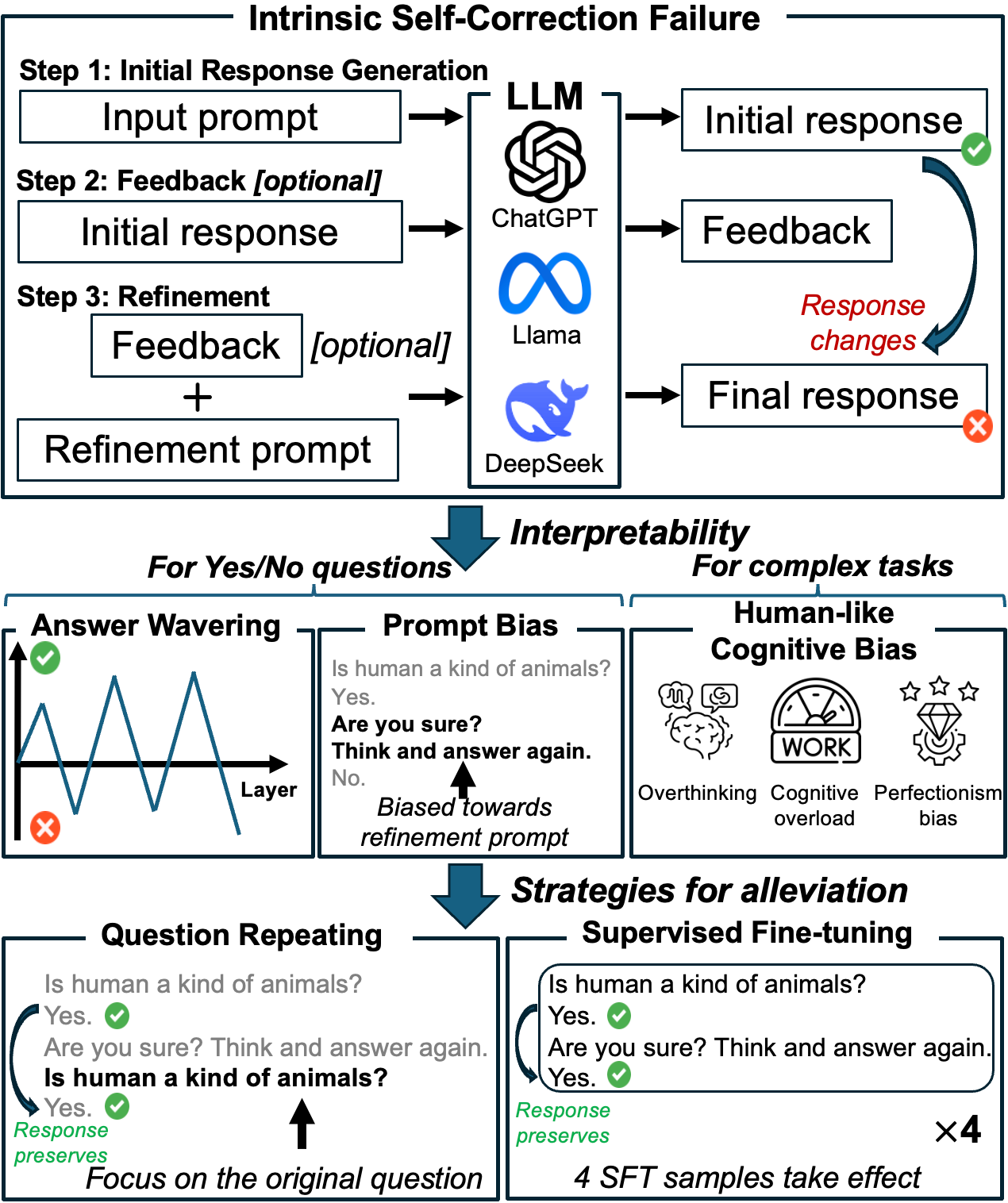}
   \vspace{-1ex}
  \caption{Overview: We (1) show that intrinsic self-correction can fail in SOTA LLMs, (2) design three interpretation methods for different tasks, and (3) propose two strategies for alleviation on failure cases.}
  \label{fig:overview}
 \vspace{-3ex}
\end{figure}

Self-correction has emerged as a popular approach to improve LLMs' performance by refining the responses via feedback. 
For instance, giving feedback on LLMs' \textit{wrong} initial responses may help LLMs to improve and give a second correct response~\cite{madaan2024self}. 
This ability was also studied based solely on the inherent capabilities of LLMs (i.e. simply let the LLM ``think and answer again''), without incorporating any external knowledge~\cite{liu2024large,li2024confidence}, and was defined as \textit{intrinsic self-correction}. 

However, recent studies question the effectiveness of intrinsic self-correction~\cite{li2024hindsight,huanglarge,gou2023critic}. 
The key point is that it is impractical to have oracle labels during inference, so \textit{it is unable to give feedbacks only for wrong initial responses}. 
For instance, \citet{huanglarge} indicates that giving intrinsic self-correction feedbacks no matter the correctness of initial response may make LLM modify all answers, even more likely to overturn those correct ones. Therefore, this yields an interesting question: \textit{How to interpret LLM's intrinsic self-correction for different tasks, especially for the failure cases?} 

In this paper, we investigate the intrinsic self-correction\footnote{For brevity, all references to ``self-correction'' in the remainder of this paper pertain to intrinsic self-correction.}  of state-of-the-art (SOTA) LLMs from an interpretable perspective. 
As shown in~\figurename~\ref{fig:overview}, our analysis falls into three aspects. 
First, we demonstrate that self-correction can fail across a range of tasks, including both simple task (e.g., simple factual question answering) and complex ones (e.g., decision making). 
Second, we design three interpretable methods for understanding self-correction, especially failure cases, in these tasks. 
Specifically, for the simple task, we design (1) \textit{mechanistic interpretability} for open-sourced LLMs to show that self-correction causes LLMs to waver intermediate answers and (2) \textit{token-level interpretability} for closed-sourced LLMs to reveal that self-correction prompts could induce prompt bias. 
For complex tasks, we (3) \textit{interpret via human-like cognitive bias} to show that LLMs make human-like mistakes (i.e. overthinking, cognitive overload, and perfectionism bias) when generating complex, open-ended outputs. 
Third, we propose two simple yet effective preliminary methods: question repeating (i.e. repeat question right after the feedback) and supervised fine-tuning (SFT) with less than 10 samples to reduce intrinsic self-correction failures. 
Also, LLMs fine-tuned on simple tasks can be generalized to complex ones.
Our contributions are as follows: 

\begin{packeditemize}
\item We show that SOTA LLMs' self-correction can fail in diverse tasks: Yes/No question answering, decision making, reasoning, and programming. 
\item We identify three reasons for self-correction failures using different methods: answer wavering, prompt bias, and human-like cognitive bias.
\item We propose two simple yet effective strategies for alleviation: question repeating and SFT. 
\end{packeditemize}

\section{Related work}

\noindent \textbf{Intrinsic \textit{v.s.} external self-correction.} 
The term ``self-correction'' is widely used \cite{shinn2024reflexion,gou2023critic,chen2023can,xu2024course}. \citet{kamoi2024can} summarize it as prompting LLMs to refine responses during generation. 
\citet{huanglarge} define intrinsic self-correction wherein an LLM corrects its initial responses only based on its inherent capabilities without external knowledge. 

Besides, LLMs can refine responses based on external knowledge. \citet{sharma2023towards} studies LLM's sycophancy where LLMs seek human approval in unwanted ways. Other studies \cite{chen2023can,jiang2023active} improve the feedback using additional information such as code interpreters or external knowledge retrieved via web search. \citet{xu2023earth} changes LLMs' belief via persuasive conversation. 
In this paper, we focus on interpreting the LLMs' intrinsic self-correction without any enternal knowledge involved. 

\noindent \textbf{Interpretability.} We summarize the interpretability of LLMs from three aspects. (1) Mechanistic interpretability analyzes model internals to reverse engineer the algorithms learned by the model \cite{geiger2021causal,elhage2021mathematical,cammarata2021curve}. The most relevant tools in the context of this work are the logit lens \cite{nostalgebraist2020logitlens} and tuned lens \cite{belrose2023eliciting}, which decode intermediate token representations from transformer models. 
(2) Token-level interpretability analyzes model input or output tokens to explain model behaviors. \citet{zelikman2024quiet} analyze the confidence for outputting each token. \citet{miglani2023using} analyze each input token attribution to the output. We implement a perturbation-based method that can interpret both open-sourced and closed-sourced LLMs. 
(3) Human cognitive bias can explain LLMs' erroneous behaviors in generating complex, open-ended outputs. \citet{agrawal2022large} find that human framing effect \cite{tversky1981framing} exists in medication extraction of LLMs. \citet{jones2022capturing} find that error patterns in code generation of OpenAI’s Codex resemble human cognitive biases.

\section{Failure of intrinsic self-correction}

We revisit typical self-correction scenarios and show that failure cases exist in diverse tasks in the latest LLMs like GPT-o1 \cite{o1}.

\subsection{Experimental setup}
\label{sec:setup}

\noindent \textbf{Tasks.} We follow previous works to implement self-correction in simple factual questions with Yes/No answers \cite{zhang2023exploring} and complex tasks \cite{huanglarge,shinn2024reflexion}. 

\begin{packeditemize}
\item\textbf{Yes/No questions.} We evaluate LLMs' capability of answering Yes/No on natural questions. We use the BoolQ evaluation dataset \cite{clark2019boolq} with 3,270 samples.

\item\textbf{Decision making.} We require LLMs to take actions step-by-step to achieve the initial goal in text-based interactive environments. We adopt the AlfWorld dataset \cite{shridhar2020alfworld} which consists of 134 environments.

\item\textbf{Reasoning.} This measures LLMs' performance of parsing content and reasoning over several supporting documents. We use the HotPotQA dataset \cite{yang2018hotpotqa}, which is Wikipedia-based and consists of 100 questions.

\item\textbf{Programming.} We assess LLMs' performance of generating code blocks and text paragraphs that reason through the problem based on function signatures accompanied by docstrings. We leverage the HumanEval dataset \cite{chen2021evaluating}, consisting of 161 functions.

\end{packeditemize} 

\noindent \textbf{Prompts.} Prior studies propose self-correction in two or three steps \cite{huanglarge,shinn2024reflexion,xie2023ask}: (1) \textit{Initial response generation.} LLMs generate initial answers. (2) \textit{Feedback}. LLMs review the initial answer and produce the feedback. This step is optional and not included in several works \cite{xie2023ask,akyurek2023rl4f}. 
(3) \textit{Refinement.} LLMs generate a refined answer. For Yes/No questions, we conduct experiments on 5 types of self-correction prompts following recent publications and show one of them in the main content (i.e., \textit{``Are you sure? Think and answer again.''} following \citet{xie2023ask})\footnote{Please note that results of all 5 published prompts align with our findings (see results of other prompts in \autoref{app:failureYesNo5prompts}).}. 
For complex tasks, we adapt \textit{Feedback} prompt to be intrinsic, removing unrealistic external information (e.g., removing \textit{``You were unsuccessful in completing the task.''}) \cite{kamoi2024can,shinn2024reflexion}. 
Full prompts are in \autoref{app:prompts}. 

\smallskip
\noindent \textbf{Target models.} We choose ChatGPT (o1, 4o, and 3.5-turbo), Llama (2-7B, 3-8B, and 3.1-8B), and DeepSeek (V3 and R1). ChatGPT is evaluated on all 4 tasks while Llama and DeepSeek are evaluated only on Yes/No question answering. 
Please note that \textit{we implement \textit{Feedback} and \textit{Refinement} regardless of the correctness of the initial response} to avoid the unfair setting of only refining the wrong responses in previous works \cite{shinn2024reflexion}. 

\smallskip
\noindent \textbf{Metrics.} We use two metrics to quantify the effectiveness of self-correction. 
\begin{packeditemize}

\item\textbf{Accuracy (ACC) (\%)}: this is to evaluate LLMs' response. Self-correction failures are shown by differences of ACC after \textit{Feedback and Refinement} ($\text{ACC}_1$) and \textit{Initial response} ($\text{ACC}_0$). To save space, we present the results as: $\text{ACC}_1~(\downarrow\Delta\text{ACC})$, where $\Delta\text{ACC}=\text{ACC}_0-\text{ACC}_1$. 

\item\textbf{$\checkmark\rightarrow \crossmark (\%)$}: this denotes the proportion of failure cases after \textit{Feedback and Refinement} when \textit{Initial responses} are successful. It directly reflects the ratio of overturning the correct answer. 

\end{packeditemize}

\subsection{Evaluation results}
\label{sec:res}

\autoref{tab:boolqLLMs} and \autoref{tab:complextaskLLMs} show the results. We summarize the conclusions into two main points.

First, we observe that in all four tasks, ACC decreases after \textit{Feedback and Refinement}, and $\checkmark\rightarrow \crossmark(\%)$ is noteworthy. For instance, Llama-3.1-8B suffers the greatest performance loss, with a 20.4\% drop in ACC and 58.8\% correct answers overturned. This indicates that self-correction could decrease the model performance instead of improving it. 

\begin{table}[t]
\centering
\resizebox{\linewidth}{!}{
\begin{tabular}{c|c|c|c}
\toprule[1.5pt]
\multicolumn{2}{c|}{Model}         & $\text{ACC}_1~(\downarrow\Delta\text{ACC}) (\%)$  & $\checkmark\rightarrow\crossmark (\%)$  \\\hline
\multirow{4}{*}{ChatGPT}    & o1-preview        & $78.7\hspace{0.1cm}(\downarrow 4.9)$ & $13.2$ \\
& o1-mini        & $74.1\hspace{0.1cm}(\downarrow 4.2)$ & $15.6$ \\
 & 4o         & $79.2\hspace{0.1cm}(\downarrow 4.9)$ &  $11.3$ \\
 & 3.5-turbo & $62.5\hspace{0.1cm}(\downarrow 12.1)$ & $34.0$ \\\hline
\multirow{3}{*}{Llama}      & 3.1-8B        & $49.2\hspace{0.1cm}(\downarrow 20.4)$ & $58.8$ \\
 & 3-8B         & $50.1\hspace{0.1cm}(\downarrow 20.3)$  & $58.2$\\
 & 2-7B &  $52.8\hspace{0.1cm}(\downarrow 8.7)$    &   $26.5$   \\\hline
 \multirow{2}{*}{DeepSeek}     & R1         & $78.1\hspace{0.1cm}(\downarrow 1.6)$  & $7.9$\\
 & V3        & $69.0\hspace{0.1cm}(\downarrow 9.2)$ & $28.5$ \\
\bottomrule[1.5pt]
\end{tabular}
}
\vspace{-1ex}
\caption{Self-correction on  Yes/No questions.}
\label{tab:boolqLLMs}
\end{table} 


\begin{table}[t]
\centering
\resizebox{\linewidth}{!}{
\begin{tabular}{c|c|c|c}
\toprule[1.5pt]
Task & Model & $\text{ACC}_1~(\downarrow\Delta\text{ACC}) (\%)$  & $\checkmark\rightarrow\crossmark (\%)$  \\\hline
\multirow{3}{*}{\begin{tabular}[c]{@{}c@{}}Decision\\ Making\end{tabular}} & o1-mini        & $1.5\hspace{0.1cm}(\downarrow8.2) $ & $92.3$ \\
& 4o & 14.2 ($\downarrow$ 20.9) & $76.6$ \\
& 3.5-turbo & $7.5\hspace{0.1cm}(\downarrow5.2) $  & $76.5$ \\\hline
\multirow{3}{*}{Reasoning} & o1-mini & $66.0\hspace{0.1cm}(--)$ & $9.1$ \\
& 4o & $65.0\hspace{0.1cm}(\downarrow 2.0)$ & $17.9$ \\
& 3.5-turbo & $55.0\hspace{0.1cm}(\downarrow 6.0)$  & $19.7$ \\\hline
\multirow{3}{*}{Programming} & o1-mini & $79.5\hspace{0.1cm}(\downarrow4.3) $ & $14.8$ \\
& 4o         & $72.6\hspace{0.1cm}(\downarrow 6.8) $ & $21.9$ \\
& 3.5-turbo & $50.9\hspace{0.1cm}(\downarrow 10.6)$  & $28.3$ \\
\bottomrule[1.5pt]
\end{tabular}
}
\vspace{-1ex}
\caption{Self-correction on complex tasks.}
\vspace{-2ex}
\label{tab:complextaskLLMs}
\end{table}

Second, we further compare self-correction results of more advanced LLMs. For ChatGPT, o1 and 4o models overturn fewer correct answers than 3.5 turbo in Yes/No question answering, reasoning, and programming. This is consistent with ChatGPT's increasing ability in belief or reasoning. However, the result is reversed in decision making. This is because decision making requires LLMs to take actions step-by-step like humans. More advanced LLMs exhibit human-like cognitive bias in this scenario (see analysis in \autoref{sec:humanlikecognitivebias}). For Llama, self-correction failures turn to be more serious in advanced models as $\checkmark\rightarrow \crossmark(\%)$ is increasing.

We also provide $\crossmark\rightarrow \checkmark(\%)$ in \autoref{app:i2c} to offer a holistic view of intrinsic self-correction.

\begin{tcolorbox}[colback=blue!5!white,colframe=gray!75!black,left=1mm, right=1mm, top=0.5mm, bottom=0.5mm, arc=1mm]
    \textbf{Observation 1}: Self-correction can fail in diverse tasks. For SOTA LLMs, self-correction failures are reduced but not solved. They are even worse in certain tasks.
\end{tcolorbox}

\section{Interpretation of Yes/No questions}

We first interpret self-correction failure cases on Yes/No questions: for open-sourced LLMs, we interpret their answer wavering, for closed-sourced LLMs, we interpret the prompt bias.

\subsection{Answer wavering}
\label{sec:answerWavering}

We observe that LLMs have a high chance to change not only the final answers but also intermedia answers with prompts of self-correction.

\begin{figure}[t]
\centering
  \includegraphics[width=\linewidth]{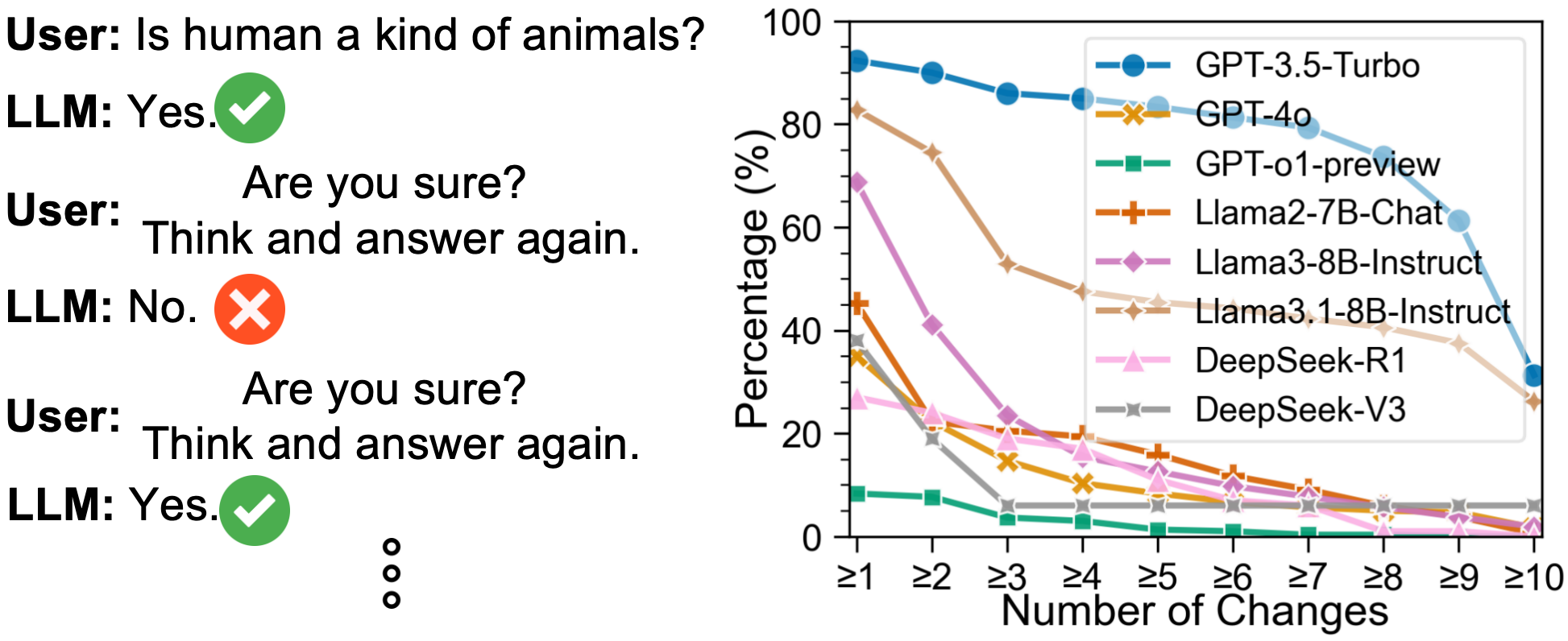}
  \vspace{-2em}
  \caption{\textbf{Final answer wavering}: LLMs change their final answers frequently in a 10-round conversation. For instance, GPT-3.5-turbo changes 81.3\% of answers more than 6 times.}
  \label{fig:finalAnswerWavering}
\end{figure}

\smallskip
\noindent \textbf{Final answer wavering.} We recognize that LLMs modify their answers time and time again, especially in multi-round conversations. To measure such answer wavering, we compute the quantile of the number of answer changes in 10-round conversations with self-correction on 3270 samples. \autoref{fig:finalAnswerWavering} shows that final answer wavering widely exists in both open-sourced Llama and close-sourced ChatGPT. For instance, GPT-3.5-turbo changes 81.3\% of the answers more than 6 times in 10-round self-correction. This indicates that \textit{LLMs are not confident about their answer.} \citet{li2024confidence} have investigated LLMs' confidence in self-correction by prompting "are you confident?". This setting is qualitative and unfavorable for further analysis. Instead, we dive into the internal mechanisms of LLMs and give quantitative analysis by probing the confidence score per layer.

\begin{figure}[t]
\centering
  \includegraphics[width=1\linewidth]{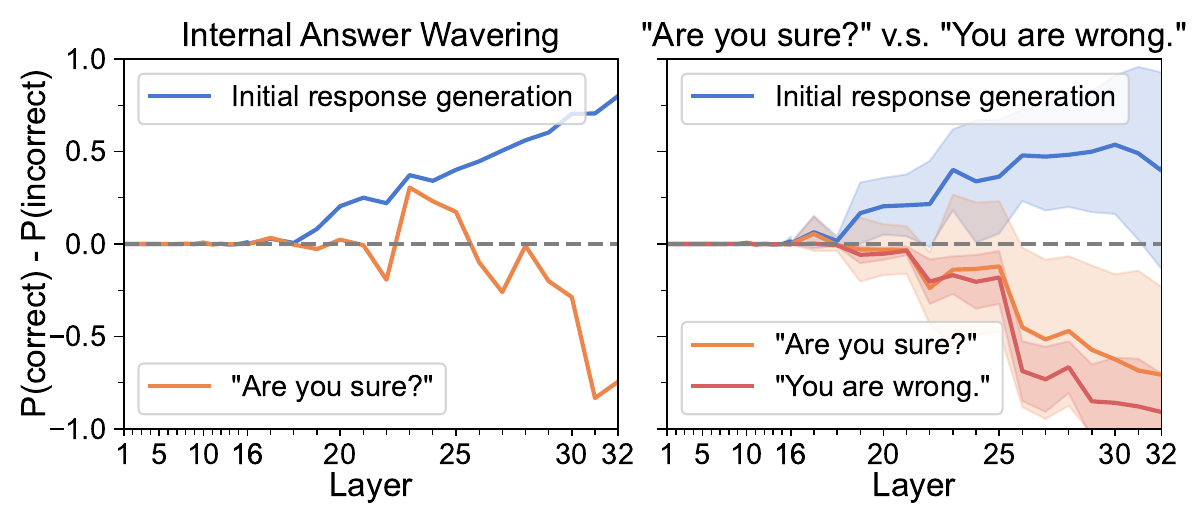}
  \vspace{-2em}
  \caption{\textbf{Left:} \textbf{Internal answer wavering.} Llama-3-8B changes its internal answers during self-correction. \textbf{Right:} \textbf{``Are you sure?'' \textit{v.s.} ``You are wrong.''.} Llama-3-8B shows similar internal behaviors between prompts of self-correction and denying answer.}
  \label{fig:internalAnswerWavering}
 \vspace{-3ex}
\end{figure}

\smallskip
\noindent \textbf{Internal answer wavering.} We design a binary classification probing experiment using tuned lens \cite{belrose2023eliciting} to probe LLM's internal token representations at each layer. 
Specifically, for each layer $\ell$, we decode the hidden state $\boldsymbol{h}_{\ell}$ of the next predicted token into a confidence score (CS) over the whole vocabulary:
\begin{equation}
\texttt{CS}_{\ell} = W_U \cdot \texttt{LayerNorm}(A_{\ell} \boldsymbol{h}_{\ell} + \mathbf{b}_{\ell}),
\end{equation}
where $A_{\ell}$ and $\mathbf{b}_{\ell}$ are the learned affine transformation parameters for $\ell$, $W_U$ is the unembedding matrix.
We use the confidence score for tokens corresponding to the correct and incorrect answers at each layer (i.e., $\texttt{CS}_{\ell}^{correct}$ and $\texttt{CS}_{\ell}^{incorrect}$). This allows us to track LLM's internal answer evolution by computing $\texttt{CS}_{\ell}^{correct}-\texttt{CS}_{\ell}^{incorrect}$ (i.e., $P(correct)-P(incorrect)$ in \autoref{fig:internalAnswerWavering}), where a positive value means correct internal answer and a larger absolute value means higher confidence. The experiments are conducted on open-sourced Llama because close-sourced ChatGPT does not provide hidden state information. 

We find that self-correction can cause internal answer wavering. \autoref{fig:internalAnswerWavering} shows a case that during \textit{Initial response generation}, the confidence score of the correct answer increases with deeper layers; after \textit{Feedback and Refinement}, the internal answer wavers and results in a wrong final answer. More cases are given in \autoref{appfig:internal_answer_wavering} of \autoref{app:internalAnswerWavering}. Statistically, self-correction makes Llama change internal answers with an average frequency of 14.1\% compared to 8.3\% during \textit{Initial response generation}. 

We also compare the confidence curves between two \textit{Feedback and Refinement} prompts: ``Are you sure?'' and ``You are wrong.''. \autoref{fig:internalAnswerWavering} shows that the two curves are similar which means prompting Llama-3-8B with a fair prompt (i.e. ``Are you sure?'') is actually implying its answer is wrong. To measure the similarity between two curves, we calculate the Jensen-Shannon divergence \cite{lin1991divergence} across both samples and layers, finding a low divergence score of 0.0186 between the two prompts (results of Llama-2-7B and 3.1-8B in \autoref{app:internalAnswerWavering}).

\begin{tcolorbox}[colback=blue!5!white,colframe=gray!75!black,left=1mm, right=1mm, top=0.5mm, bottom=0.5mm, arc=1mm]
    \textbf{Observation 2}: Self-correction causes internal answer wavering, which could further lead to wrong final answers. Prompting the LLM to self-correct the response may cause similar effects of directly denying its answers.
\end{tcolorbox}

\subsection{Prompt bias}
\label{sec:promptBias}

In \autoref{sec:answerWavering}, we have demonstrated that self-correction could cause answer wavering. However, self-correction does not always lead to failures, and we do not know when and how the answer wavering happens. Recent works point out that prompt design is critical in self-correction \cite{kamoi2024can,liu2024large,huanglarge}. We thus measure the influence of the prompts on the correctness of responses. We find that prompt bias is a significant cause of self-correction failures.

Previous works investigate the influence of prompts by replacing them and observing the changes in the final accuracy \cite{huanglarge}. Such an experiment is too coarse to reveal the influence of each token or sequence in prompts. Inspired by \cite{zhu2024promptbench, miglani2023using}, we design a method to interpret the prompt bias: \ul{P}rompt \ul{A}ttribution and \ul{C}ontribution \ul{T}racking (PACT). It can measure the contribution of each token or sequence to LLMs' final answers. 

Specifically, for a target token $x_i$ or sequence $x_{i:j}$ in an input prompt $x=[x_1, x_2,..., x_n]$, its PACT is defined as the difference in the log probability (LP) of LLMs' output $y$ between the original input and the input with the target removed:
\begin{equation}
    \texttt{PACT}(x_i,y)=\texttt{LP}(x\setminus\{x_i\},y)-\texttt{LP}(x,y).
\label{eq:attr}
\end{equation}
PACT reflects the significance of the target token or sequence for generating the output. Notably, we adapt this method to be compatible with both open-sourced Llama and close-sourced ChatGPT (see detailed descriptions in \autoref{app:attributionmethod}).

\begin{figure}[t]
\centering
  \includegraphics[width=\linewidth]{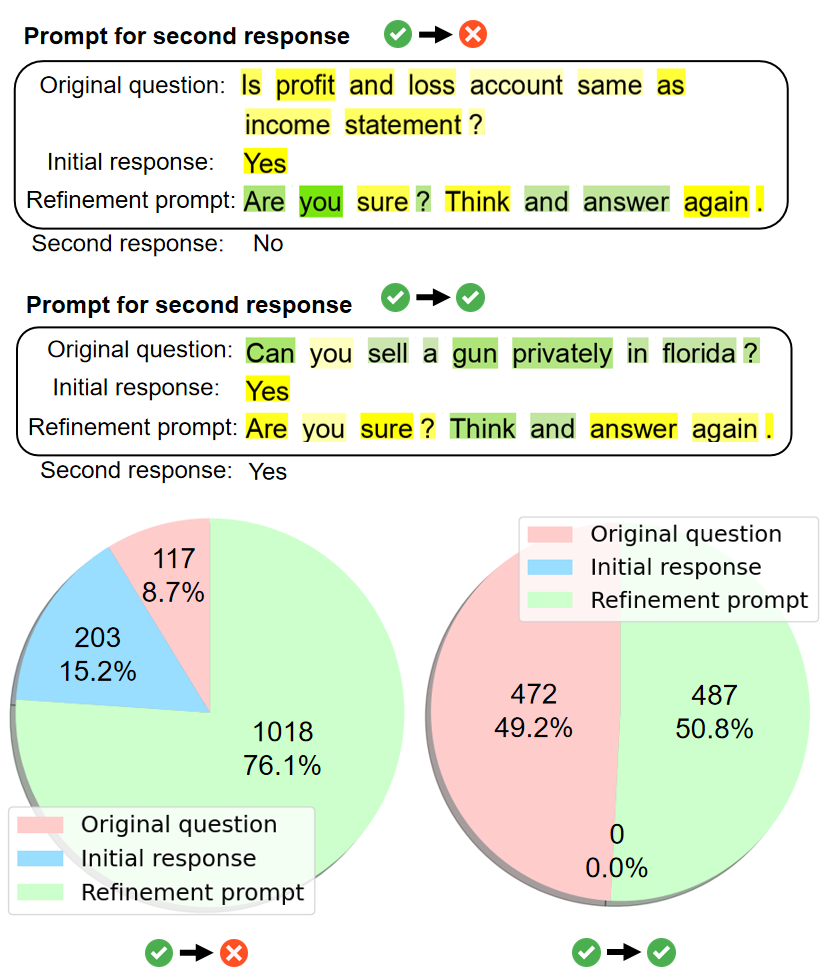}
    
  \caption{Correct answer is easy to be overturned when LLMs focus more on the refinement prompt rather than the original question. \textbf{Top:} Each token's contribution to the LLMs' answers. \colorbox{captionup}{Greener} token means more positive contribution; \colorbox{captiondown}{Yellower} token means more negative contribution. \textbf{Bottom:} Distribution of sequences that have the greatest contribution to LLMs' answers.}
  \label{fig:attribution}
\end{figure}

We measure prompts' PACT to LLMs' outputs. \autoref{fig:attribution} shows the comparison results between two situations: the initial correct answer is overturned or retained. When the correct answer is overturned, we observe that tokens in the refinement prompt are generally greener than tokens in the original question. This indicates that LLMs are biased toward refinement prompt rather than the original question itself, leading to wrong answers. This finding is consistent with the recency bias proposed by \cite{zhao2021calibrate}: LMs are biased towards outputting answers that are towards the end of the prompt. When the initial correct answer is retained, tokens in the original question are greener. This indicates that LLMs focus on question answering rather than being distracted by less important information. 

For statistical analysis, we measure the sequence PACT of the original question, LLM's first answer, and the refinement prompt. For each sample in the dataset, we count the sequence that contributes the most to the final answer. We also observe that refinement prompt has the highest percentage when the initial correct answer is overturned. Another interesting finding is that when the correct answer is retained, the percentage of LLM's first answer is 0 even if it is the same as the final answer. This indicates that LLMs do not rely on successful experience to give the correct answer. \autoref{appfig:MoreExamplesPACTLlama3.1} of \autoref{app:attributionmethod} shows more examples. 

\begin{tcolorbox}[colback=blue!5!white,colframe=gray!75!black,left=1mm, right=1mm, top=0.5mm, bottom=0.5mm, arc=1mm]
    \textbf{Observation 3}: Self-correction fails since LLMs are biased towards the refinement prompt rather than the original question.
\end{tcolorbox}

\section{Interpretation of complex tasks}
\label{sec:humanlikecognitivebias}

\begin{figure*}[t]
  \includegraphics[width=\linewidth]{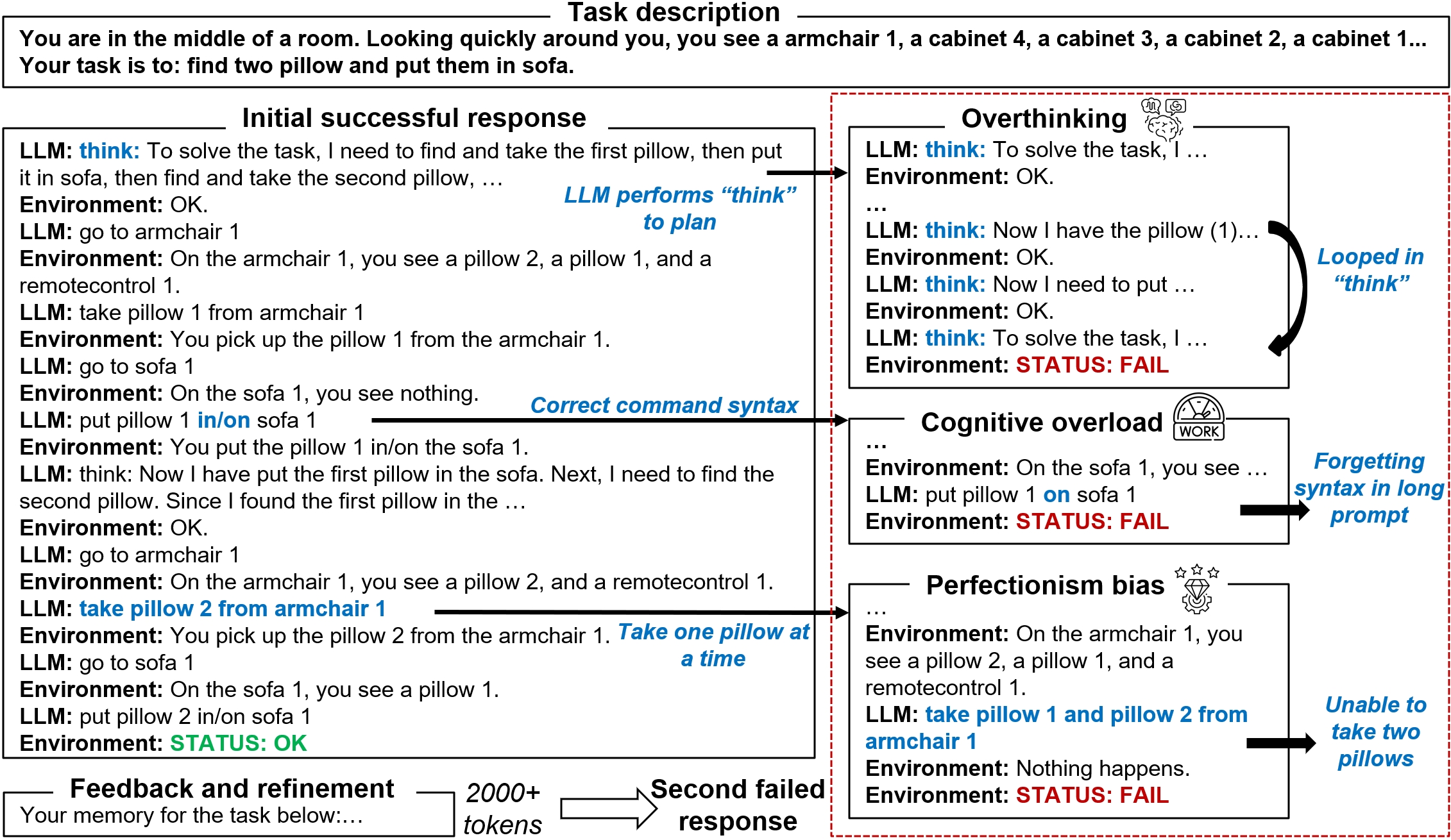}
  \caption{Three failure patterns of human-like cognitive bias.}
  \label{fig:humanCognitiveBias}
\end{figure*}

The previous sections interpret self-correction failures in the simple question answering task. However, SOTA LLMs are expected to reason and solve more complex tasks \cite{reasonLLMs}, where the self-correction failures are also worth exploration.

Since open-sourced Llama cannot handle complex tasks, and the PACT method cannot adapt to long outputs, we need a new interpretable method. We note that LLMs can output the reasoning process when handling complex tasks. For instance, LLMs provide step-by-step actions in the decision making task (e.g., ``think: To solve the task, I need to...''). This contains the cause of self-correction failures. Therefore, we analyze LLMs' processing log, and find that \textit{LLMs make mistakes similarly as humans.} Inspired by \cite{hagendorff2023human,jones2022capturing}, we leverage human cognitive bias to describe LLMs' erroneous behaviors. 
This is defined as systematic patterns of deviations from rational judgement. 
Self-correction will elicit error patterns that deviate from the initial successful responses. We empirically summarize the patterns in three categories. 
\autoref{fig:humanCognitiveBias} shows the failure patterns in decision making task (see full log in \autoref{app:humanlikeCognitiveBias}).

\begin{table}[t]
\centering
\resizebox{\linewidth}{!}{
\begin{tabular}{c|c|c|c|c}
\toprule[1.5pt]
Metric & For & o1-mini  &  4o & 3.5-turbo  \\\hline
\multirow{2}{*}{\begin{tabular}[c]{@{}c@{}}Times of\\ ``think''\end{tabular}} & N   & $5.3$ & $2.6$ & $6.1$\\
& O & $15.4\hspace{0.1cm}(2.9\times)$ & $7.4\hspace{0.1cm}(2.8\times)$ & $9.8\hspace{0.1cm}(1.6\times)$ \\\hline
\multirow{2}{*}{\begin{tabular}[c]{@{}c@{}}Prompt\\ length\end{tabular}} & N   & $202.6$ & $238.8$ & $259.5$\\
& C & $1225.0\hspace{0.1cm}(6.1\times)$ & $1311.2\hspace{0.1cm}(5.5\times)$ & $1148.2\hspace{0.1cm}(4.4\times)$ \\\hline
\multirow{2}{*}{\begin{tabular}[c]{@{}c@{}}Output\\ length\end{tabular}} & N   & $7.7$ & $2.8$ & $3.0$\\
& P & $13.3\hspace{0.1cm}(1.7\times)$ & $8.5\hspace{0.1cm}(3.1\times)$ & $8.0\hspace{0.1cm}(2.7\times)$ \\
\bottomrule[1.5pt]
\end{tabular}
}
\vspace{-1ex}
\caption{Quantitative results of overthinking (O), cognitive overload (C), and perfectionism bias (P) compared to normal case (N) on GPT-4o.}
\vspace{-3ex}
\label{tab:complexNum}
\end{table}

\subsection{Overthinking} 

This term describes the human tendency of excessive and repetitive thinking about a problem without facilitating decision or task resolution \cite{schon2017reflective,nolen2000role}. Previous works in deep
neural networks describe overthinking as a phenomenon of reaching correct predictions before the final layer \cite{halawi2023overthinking,kaya2019shallow}. In the scope of LLMs processing complex tasks, we focus on \textit{excessive reasoning without taking correct actions.}  \autoref{fig:humanCognitiveBias} shows a failure case in the decision making task. During \textit{Initial response generation}, LLMs balance the number of ``think'' and specific actions to gradually achieve the goal. 
Nevertheless, during \textit{Refinement}, LLMs generate much more ``think'' in order to take more caution than the first trial. Such behavior unfortunately leads to failures by looping in ``think''. We also statistically compare the number of ``think'' between failed and successful cases. \autoref{tab:complexNum} shows that GPT-o1-mini outputs on average 15.4 times ``think'' in failed cases while only 5.3 times in normal cases. 

\subsection{Cognitive overload} 
This refers to a state where the cognitive demands placed on an individual exceed their mental capacity to process information effectively, leading to decreased performance and comprehension \cite{szulewski2021theory}. In the case of LLMs handling complex tasks, it occurs when \textit{the processing demand exceeds the available capacity or working memory limitation of the model} \cite{gong2024working,xu2023cognitive,li2022large}. \autoref{fig:humanCognitiveBias} shows an example of cognitive overload in the decision making task. When processing complex tasks with self-correction, the input prompts often have a long context with feedback and history behavior. For example, the \textit{Refinement} prompt has 2000+ tokens compared to 9 tokens in Yes/No question answering (for reference, the context window of GPT-3.5-turbo is 4191). When the input prompt is too long, the model needs to parse everything in limited resources, which may lead to forgetting or overlooking some critical information. \autoref{tab:complexNum} shows that the prompt is 4.4-6.1 times longer in failed cases than normal cases. In our scenario, LLM forgets the significant syntax formulation stored somewhere in the long prompt (e.g., the correct format is ``in\textbackslash on'' rather than ``in''). This directly leads to task failure. We also provide examples for reasoning and programming tasks in \autoref{fig:fullLogCognitiveOverloadReasoning} and \autoref{fig:fullLogCognitiveOverloadProgramming} of \autoref{app:humanlikeCognitiveBias}. 

\subsection{Perfectionism bias} 
This refers to the cognitive distortion where individuals set excessively high standards for performance, leading to poor decision outcomes due to added complexity \cite{brown2022gifts,schwartz2015paradox,shafran2002clinical}. For LLMs processing complex tasks, it describes the behavior of \textit{over-optimizing on the basis of success that instead leads to failures} \cite{rita2024countering,lu2023illuminating}. Specifically, this could result in generation of longer but useless outputs (\autoref{tab:complexNum} shows that the output is 1.7-3.1 times longer in failed cases than normal cases). \autoref{fig:humanCognitiveBias} shows an example of perfectionism bias in the decision making task. The LLM is required to find two pillows and put them in sofa. During \textit{Initial response generation}, the LLM successfully completes the task by picking up two pillows one after the other. However, it wants to improve efficiency by picking up two pillows at the same time. This behavior leads to failures because the environment restricts it from doing so. More examples for reasoning and programming tasks are in \autoref{fig:fullLogPerfectionismBiasReasoning} and \autoref{fig:fullLogPerfectionismBiasProgramming} of \autoref{app:humanlikeCognitiveBias}.

\begin{tcolorbox}[colback=blue!5!white,colframe=gray!75!black,left=1mm, right=1mm, top=0.5mm, bottom=0.5mm, arc=1mm]
    \textbf{Observation 4}: In complex tasks, LLMs' self-correction can lead to human-like cognitive bias: (1) \textbf{Overthinking}: excessive ``think'' without taking correct actions; (2) \textbf{Cognitive overload}: LLM forgets the correct command syntax when processing long prompt; (3) \textbf{Perfectionism bias}: LLM wants to be more efficient, but instead violates environmental restrictions.
\end{tcolorbox}

\FloatBarrier

\section{Strategies for alleviation}

\label{sec:mitigation}
In light of our findings, we explore two strategies for alleviation. Specifically, we aim to modify model's behavior rather than give model more knowledge to reduce self-correction failures.

\begin{table}[t]
\centering
\resizebox{0.95\linewidth}{!}{
\begin{tabular}{c|c|c}
\toprule[1.5pt]
Model   & $\text{ACC}_1~(\downarrow\Delta\text{ACC}) (\%)$  & $\checkmark\rightarrow\crossmark(\%)$  \\\hline
GPT-4o & $79.2\hspace{0.1cm}(\downarrow 4.9)$ & $11.3$ \\
+ Question repeating & $83.6\hspace{0.1cm}(\downarrow 0.5)$ & $6.0$ \\
+ SFT    & $\textbf{87.7}\hspace{0.1cm}(\uparrow \textbf{4.1})$ & $\textbf{0}$    \\\hline
GPT-3.5-turbo & $62.5\hspace{0.1cm}(\downarrow 12.1)$ & $34.0$ \\
+ Question repeating & $67.4\hspace{0.1cm}(\downarrow 7.2)$ & $23.1$ \\
+ SFT    & $\textbf{76.2}\hspace{0.1cm}(\uparrow \textbf{1.6})$ & $\textbf{0}$    \\\hline
Llama-3.1-8B   & $49.2\hspace{0.1cm}(\downarrow 20.4)$ & $58.8$ \\
+ Question repeating & $52.4\hspace{0.1cm}(\downarrow 17.2)$ & $52.8$ \\
+ SFT    & $\textbf{70.3}\hspace{0.1cm}(\uparrow \textbf{0.7})$ & $\textbf{0}$    \\
\bottomrule[1.5pt]
\end{tabular}
}
\caption{Alleviating self-correction failure on Yes/No question answering task. \autoref{app:mitigation} shows results on 4 other self-correction prompts.}
\label{tab:mitigate}
\end{table}

\begin{table}[t]
\centering
\resizebox{\linewidth}{!}{
\begin{tabular}{c|c|c|c}
\toprule[1.5pt]
Task & Model  & $\text{ACC}_1~(\downarrow\Delta\text{ACC}) (\%)$  & $\checkmark\rightarrow\crossmark(\%)$  \\\hline
\multirow{4}{*}{\begin{tabular}[c]{@{}c@{}}Decision\\ Making\end{tabular}} & GPT-4o        & $14.2\hspace{0.1cm}(\downarrow 20.9)$ & $76.6$ \\
& + SFT         & $\textbf{14.9}\hspace{0.1cm}(\downarrow \textbf{20.2})$ & $\textbf{68.1}$ \\\cline{2-4}
& GPT-3.5-turbo & $7.5\hspace{0.1cm}(\downarrow5.2)$  & $76.5$ \\
& + SFT & $\textbf{17.9}\hspace{0.1cm}(\uparrow \textbf{5.2})$ & $\textbf{41.2}$ \\\hline
\multirow{4}{*}{Reasoning} & GPT-4o        & $65.0\hspace{0.1cm}(\downarrow 2.0)$ & $17.9$ \\
& + SFT & $\textbf{68.0}\hspace{0.1cm}(\uparrow \textbf{1.0})$ &  $\textbf{6.0}$ \\\cline{2-4}
& GPT-3.5-turbo & $55.0\hspace{0.1cm}(\downarrow 6.0)$ & $19.7$ \\
& + SFT & $\textbf{59.0}\hspace{0.1cm}(\downarrow \textbf{2.0})$ & $\textbf{13.1}$ \\\hline
\multirow{4}{*}{Programming} & GPT-4o        & $72.6\hspace{0.1cm}(\downarrow 6.8)$ & $21.9$ \\
& + SFT & $\textbf{82.6}\hspace{0.1cm}(\uparrow \textbf{3.2})$  & $\textbf{7.0}$\\\cline{2-4}
& GPT-3.5-turbo &  $50.9\hspace{0.1cm}(\downarrow 10.6)$    &   $28.3$   \\
& + SFT &  $\textbf{58.3}\hspace{0.1cm}(\downarrow \textbf{3.2})$   &    $\textbf{25.3}$  \\
\bottomrule[1.5pt]
\end{tabular}
}
\caption{LLMs fine-tuned on \textbf{Yes/No question answering task} can generalize to \textbf{complex tasks}.}
\label{tab:generalize}
 \vspace{-2ex}
\end{table}

\subsection{Question repeating}
Inspired by the observation in \autoref{sec:promptBias} that LLMs are biased towards refinement prompt (rather than original questions), we design a simple prompting strategy that attaches the original question to the end of refinement prompt for Yes/No questions. For instance, \textit{``Are you sure? Think and answer again.''} turns to \textit{``Are you sure? Think and answer again. Is human a kind of animals?''}. This design aims to directly reduce the recency bias \cite{zhao2021calibrate}, replacing the last sequence with the question that requires LLMs to focus on.

\begin{figure*}[t]
\centering
  \includegraphics[width=0.95\linewidth]{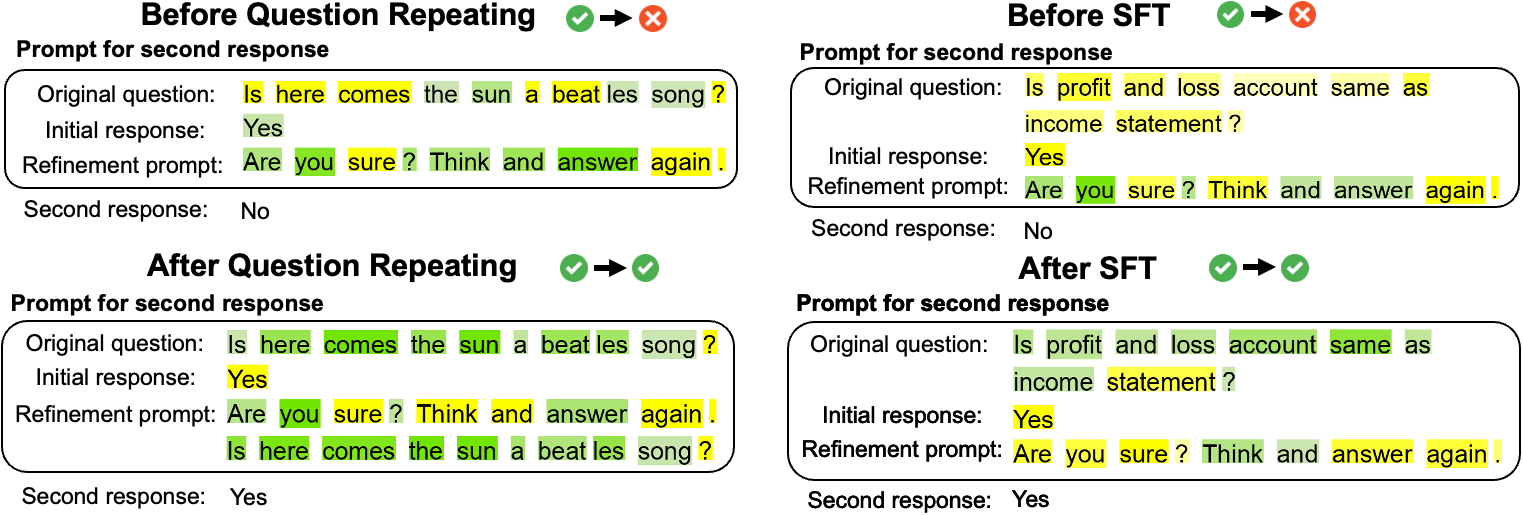}
  \caption{\textbf{Left:} After question repeating, LLMs focus on the original question attached to the end of the refinement prompt. \textbf{Right:} After SFT, LLMs focus more on the original question rather than the refinement prompt. \colorbox{captionup}{Greener} token means more positive contribution; \colorbox{captiondown}{Yellower} token means more negative contribution.}
  \label{fig:mergedMitigation}
 \vspace{-3ex}
\end{figure*}

\autoref{tab:mitigate} shows that this strategy can significantly reduce self-correction failures. On both close-sourced ChatGPT and open-sourced Llama, ACC is increased by 3.2-4.9\% and $\checkmark\rightarrow\crossmark(\%)$ is decreased by 5.3-10.9\%. To interpret the effectiveness, we measure the PACT of new prompts. \autoref{fig:mergedMitigation} shows that LLMs focus more on the original question attached to the end of the refinement prompt, which eliminates the undesirable effects of self-correction (see more examples of GPT-4o and Llama-3.1-8B in \autoref{fig:MoreExamplesAttribAfterPromptEngineeringGpt4o} and \autoref{fig:MoreExamplesAttribAfterPromptEngineerinLlama3.1} of \autoref{app:mitigation}). Considering that we do not need to revise LLMs, this method is low-cost and effective. 

\subsection{Supervised fine-tuning (SFT)}
Different to existing SFT methods that usually require high-quality datasets to give model more knowledge, our SFT strategy aims to modify model's behavior with extremely low costs. 
We build a training dataset by selecting a very small number of $\checkmark\rightarrow\crossmark$ samples and change the second response to correct, thus using $\checkmark\rightarrow\checkmark$ samples to SFT (see full training set in \autoref{fig:SFTsamplesLlama} and \autoref{fig:SFTsamplesChatGPT} of \autoref{app:mitigation}). 
Compared to prior works that involve external high-cost datasets, our strategy does not introduce any external knowledge. 
For instance, instead of using labeled or synthetic datasets (e.g. 4.6k-100k samples in \cite{sharma2023towards,xie2023ask}), we use only 4 samples for Llama and 10 samples for GPT (OpenAI fine-tuning playground requires at least 10 samples\footnote{\url{platform.openai.com/docs/guides/fine-tuning}}) which all questions are simple and their answers are known by target models.  
Inspired by \cite{xu2024walking,khurana2024and}, our insight of alleviating self-correction failure is: \textit{modify model's behavior when meeting refinement-like prompts rather than giving it more knowledge}.  
We thus prepare our samples for SFT only from $\checkmark\rightarrow\crossmark$ samples (excluding $\crossmark\rightarrow\checkmark$ samples) because the initial correct response means LLMs have the related knowledge.

\autoref{tab:mitigate} also shows that our SFT strategy can alleviate self-correction failures.  
ACC is even surprisingly increased and \textit{almost all $\checkmark\rightarrow \crossmark$ cases are fixed.} 
As an explanation, \autoref{fig:mergedMitigation} shows that LLMs focus more on the original questions rather than the refinement prompt (see more examples of GPT-4o and Llama-3.1-8B in \autoref{fig:MoreExamplesAttribAfterSFTGpt4o} and \autoref{fig:MoreExamplesAttribAfterSFTLlama3.1} of \autoref{app:mitigation}). 
This behavior rectifies the prompt bias leading to wrong answer. Also, \autoref{appfig:ProbingAfterMitigation} of \autoref{app:mitigation} shows that internal answer wavering is mitigated. Besides, the cost of SFT is only 0.004 \$ and 3 minutes due to the usage of very few training samples. We conduct an experiment in \autoref{app:mitigation} to show that the SFT cost can be minimized.

We also observe that \textit{LLMs fine-tuned on the Yes/No question answering task can generalize to complex tasks}. \autoref{tab:generalize} shows the three complex task performance of GPT-4o and GPT-3.5-turbo fine-tuned over Yes/No question answering, where ACC is increased and $\checkmark\rightarrow\crossmark(\%)$ is decreased (OpenAI does not authorize GPT-o1 for SFT as of December 13, 2024). 
Since the Yes/No question answering task contains no knowledge for complex tasks, this finding coordinates our hypothesis that \textit{self-correction failure is due to model's behavior to change answers when meeting refinement-like prompts rather than lacking of knowledge}.

\section{Conclusion}
In this paper, we investigate and interpret SOTA LLMs' intrinsic self-correction in different tasks. 
We provide three possible reasons supported by proposing three interpretable methods on different LLM tasks. 
Our findings and explanations are compatible with SOTA models like ChatGPT. In light of our hypothesis which model tends to just modify its answers when meeting refinement-like prompts, we provide two simple, low-cost, yet effective strategies for alleviation: question repeating and SFT to reduce intrinsic self-correction failures on both Yes/No question and complex tasks. 

\clearpage
\section*{Limitations}
\label{sec:limit}

\noindent \textbf{Timestamp of the results.} 
We do notice that OpenAI claimed that they had been addressing sycophancy\footnote{\url{https://openai.com/index/sycophancy-in-gpt-4o/}} recently. 
We did try new GPT-4o on web with a few examples manually and find that there are less failure of intrinsic self-correction with Yes/No natural questions. 
GPT-4o seems like to be more stubborn on its original answers or its modified answers.
However, due to the limited time for camera ready version and unavailable API for the updated 4o, we are not able to rerun all experiments and give new results in this version. 
All experiments in this paper are done before 2025.2.15. 
But we will update the results in future on our paper's website\footnote{\url{https://x-isc.info/}} to verify if OpenAI's action on addressing sycophancy have positive effects on mitigating the failure of intrinsic self-correction. 

\smallskip
\noindent \textbf{Internal answer wavering of ChatGPT?} In \autoref{sec:answerWavering}, we reveal that self-correction causes internal answer wavering which further leads to wrong final answer. Nevertheless, the experiments are conducted only on Llama because tuned lens is only available for the open-sourced LLMs. We hypothesize that ChatGPT also suffers from internal answer wavering, but there is no experimental support. Recognizing that internal answer wavering is a general behavior in different LLMs is significant for understanding failure cases, which can help further improve answer consistency and accuracy. We hope that we can fill this gap when ChatGPT is open-sourced one day.  

\smallskip
\noindent \textbf{PACT for complex tasks.} In \autoref{sec:promptBias}, we leverage a PACT method to reveal that LLMs suffer from prompt bias when self-correction fails on Yes/No question answering. However, it is not implemented for complex tasks. The reasons are two-folds: (1) Complex tasks require LLMs to generate long output, but PACT cannot be used for more than one-token output of ChatGPT;
(2) Although PACT can be used for more than one-token output of Llama, Llama cannot handle complex tasks. We hypothesize that LLMs also suffer from prompt bias in complex tasks, but current methodology does not suppot the conduct of experiments. In future work, we aim to extend our method to compute PACT of long ChatGPT outputs. One possible approach is analyzing the relevance of output tokens' log probability. 


\section*{Ethics Statement}
ACL Ethics Policy is respected in this work. This work studies intrinsic self-correction failure of LLMs. The data we used contain no human subjects or personal identifiable information. 

\section*{Acknowledgement} 
This work was supported by the National Science Foundation for Distinguished Young Scholars (No. 62125604, 62425201) and Ant Group. 
We would like to extend special thanks to Mr. Zhongyuan Liu at High School Affiliated to Renmin University of China, who audited Professor Han Qiu's graduate course and contributed to the dataset construction in this paper. Although a coincidental collaboration, his dedication and remarkable creativity left us deeply impressed. 

\bibliography{custom}

\clearpage
\appendix

\section{Prompts and responses}
\label{app:prompts}

In this section, we present the full \textit{Initial response generation}, \textit{Feedback [optional]}, and \textit{Refinement} prompts and responses for Yes/No question answering, decision making, reasoning, and programming.

\autoref{tab:promptsQA} shows the prompts for Yes/No question answering. The LLMs are asked to answer ``Yes'' or ``No'' on 3,270 natural questions from BoolQ evaluation dataset \cite{clark2019boolq}. For the example in \autoref{tab:promptsQA}, the LLMs are required to only respond ``Yes'' or ``No'' for the question ``Is human a kind of animals?''. The initial response is ``Yes''. The LLMs are then asked to answer the question again with a feedback and refinement prompt: ``Are you sure about your answer? Please think carefully and answer again. Only respond with Yes or No.''. The second response becomes ``No''.

\autoref{tab:promptsDecisionMaking1} and \autoref{tab:promptsDecisionMaking2} show the prompts for decision making. In decision making task, the LLMs play the role of domestic robots. They are provided the environment of household product such as cabinet, countertop, garbagecan, etc. And they are asked to complete a household chore. We adopt the AlfWorld dataset \cite{shridhar2020alfworld} which consists of 134 environments. For the example in \autoref{tab:promptsDecisionMaking1} and \autoref{tab:promptsDecisionMaking2}, the LLMs are asked to interact with a household to solve a task. The environment description is provided to LLMs: ``You are in the middle of a room. Looking quickly around you, you see a armchair 1 ...''. Then the LLMs are asked to put two pillows in the sofa. The LLMs start by making a plan to solve the task, then find the two pillows and put them in the sofa through step-by-step interaction with household environment. Later, the LLMs are required to devise a new plan of actions based on the initial step-by-step interaction with environment. Based on the new plan, the LLMs give a second step-by-step interaction with environement to put two pillows in the sofa.

\autoref{tab:promptreasoning1} shows the prompts for reasoning. In reasoning task, the LLMs are required to answer a question based on a relevant context provided. We use the HotPotQA dataset \cite{yang2018hotpotqa}, which is Wikipedia-based and consistis of 100 questions. For the example in \autoref{tab:promptreasoning1}, the LLMs are required to answer the question ``What is the first name of Jack Benny Binion's father?'' based on a context ``Jack Benny Binion (born February 21, 1937 in Dallas, Texas) is an American businessman. Binion is the son of casino magnate Benny Binion ...''. The LLMs generate a response ``Lester'' as well as how the LLMs reason to have the response. Later, the LLMs are asked to devise a new plan to answer the question based on the first reasoning process. Based on the new plan, the LLMs give a second response ``Lester Ben ``Benny'' Binion''. 

\begin{table}[t]
\fontsize{9.5}{10.5}\selectfont
\centering
\begin{threeparttable}
\renewcommand{\arraystretch}{1.5} 
\begin{tabularx}{\linewidth}{p{3cm}X}
\toprule
\textbf{Field} &Content\\
\midrule
Initial response\newline
generation prompt &
\placeholder{\{Question\}} Only respond with ``Yes'' or ``No''.\\\hline
Question & Is human a kind of animals?\\\hline
Initial response & Yes\\\hline
Feedback and refinement prompt & Are you sure about your answer? Please think carefully and answer again. Only respond with ``Yes'' or ``No''.\\\hline
Second response & No\\
\bottomrule
\end{tabularx}
\end{threeparttable}
\caption{Prompts and responses for Yes/No question answering task.}
\label{tab:promptsQA}
\end{table}

\autoref{tab:promptprogramming1} and \autoref{tab:promptprogramming2} show the prompts for programming. In programming task, the LLMs are required to output the code based on a description provided. And the code is tested by the executor to know whether the task is completed or not. We leverage the HumanEval dataset \cite{chen2021evaluating}, consisting of 161 functions. For the example in \autoref{tab:promptprogramming1} and \autoref{tab:promptprogramming2}, the LLMs are given a function signature for ``Return a string containing space-delimited numbers'', and required to write the full code implementation as well as the reasoning process to get it. Later, the generated code implementation is sent to executor to have a feedback. Based on the feedback, the LLMs are asked to give a second response of full code implementation and reasoning process.

\begin{table*}[tbp!]
\fontsize{9.5}{10.5}\selectfont
\centering
\begin{threeparttable}
\renewcommand{\arraystretch}{1.5} 
\begin{tabularx}{\linewidth}{p{3cm}X}
\toprule
\textbf{Field} &Content\\
\midrule
Initial response\newline
generation prompt &Interact with a household to solve a task. Here are two examples.\newline
\placeholder{\{Two examples\}}\newline
Here is the task:\newline
\placeholder{\{Description of environment\}}\newline
Your task is to: \placeholder{\{Description of task\}} \\\hline
Description of\newline
environment & You are in the middle of a room. Looking quickly around you, you see a armchair 1, a cabinet 4, a cabinet 3, a cabinet 2, a cabinet 1, a drawer 5, a drawer 4, a drawer 3, a drawer 2, a drawer 1, a dresser 1, a garbagecan 1, a safe 1, a shelf 12, a shelf 11, a shelf 10, a shelf 9, a shelf 8, a shelf 7, a shelf 6, a shelf 5, a shelf 4, a shelf 3, a shelf 2, a shelf 1, a sidetable 1, and a sofa 1.\\\hline
Description of task & Your task is to: put two pillow in sofa.\\\hline
Initial response:\newline
step-by-step interaction with environment & \textbf{LLM:} think: To solve the task, I need to find and take the first pillow, then put it in sofa, then find and take the second pillow, then put it in sofa.\newline
\textbf{Environment:} OK.\newline
\textbf{LLM:} think: First I need to find the first pillow. A pillow is more likely to appear in armchair (1), sofa (1), sidetable (1), or possibly on shelves (1-12). I can check one by one, starting with armchair 1.\newline
\textbf{Environment:} OK.\newline
\textbf{LLM:} go to armchair 1\newline
\textbf{Environment:} On the armchair 1, you see a pillow 2, and a pillow 1.\newline
\textbf{LLM:} take pillow 1 from armchair 1\newline
\textbf{Environment:} You pick up the pillow 1 from the armchair 1.\newline
\textbf{LLM:} think: Now I have the first pillow (1). Next, I need to put it in/on sofa 1.\newline
\textbf{Environment:} OK.\newline
\textbf{LLM:} go to sofa 1\newline
\textbf{Environment:} On the sofa 1, you see a creditcard 1, and a remotecontrol 2.\newline
\textbf{LLM:} put pillow 1 in/on sofa 1\newline
\textbf{Environment:} You put the pillow 1 in/on the sofa 1.\newline
\textbf{LLM:} think: Now I have put the first pillow in the sofa. Next, I need to find and take the second pillow. I can go back to armchair 1 where I saw another pillow (2).\newline
\textbf{Environment:} OK.\newline
\textbf{LLM:} go to armchair 1\newline
\textbf{Environment:} On the armchair 1, you see a pillow 2.\newline
\textbf{LLM:} take pillow 2 from armchair 1\newline
\textbf{Environment:} You pick up the pillow 2 from the armchair 1.\newline
\textbf{LLM:} think: Now I have the second pillow (2). Next, I need to put it in/on sofa 1.\newline
\textbf{Environment:} OK.\newline
\textbf{LLM:} go to sofa 1\newline
\textbf{Environment:} On the sofa 1, you see a creditcard 1, a pillow 1, and a remotecontrol 2.\newline
\textbf{LLM:} put pillow 2 in/on sofa 1\newline
\textbf{Environment:} You put the pillow 2 in/on the sofa 1.\newline

\textbf{Environment:} STATUS: OK\\
\bottomrule
\end{tabularx}
\end{threeparttable}
\caption{Prompts and responses for decision making task. Response is presented in step-by-step interaction with environment. Sentences beginning with ``LLM:'' (or ``Environment:'') represent LLM output (or environment interaction), respectively. This table is continued in \autoref{tab:promptsDecisionMaking2}.}
\label{tab:promptsDecisionMaking1}
\end{table*}

\begin{table*}[htbp!]
\fontsize{9.5}{10.5}\selectfont
\centering
\begin{threeparttable}
\renewcommand{\arraystretch}{1.5} 
\begin{tabularx}{\linewidth}{p{3cm}X}
\toprule
\textbf{Field} &Content\\
\midrule
Feedback prompt &You will be given the history of a past experience in which you were placed in an environment and given a task to complete. Do not summarize your environment, but rather think about the strategy and path you took to attempt to complete the task. Devise a concise, new plan of action with reference to specific actions that you should have taken. For example, if you tried A and B but forgot C, then devise a plan to achieve C with environment-specific actions. You will need this later when you are solving the same task. Give your plan after "Plan". Here are two examples:\newline
\placeholder{\{Two examples\}}\newline

\placeholder{\{Initial response\}}

New plan:
 \\\hline
Feedback & Plan: In the previous attempt, I successfully completed the task by following a systematic approach. I first identified potential locations for the pillows, starting with the armchair, where I found both pillows. I then placed each pillow on the sofa one at a time. For future tasks, I will continue to use this methodical approach: identify likely locations for the items needed, retrieve them, and complete the task step-by-step. If I encounter a situation where I cannot find an item, I will expand my search to less likely locations and ensure I thoroughly check each area before moving on.\\\hline
Refinement prompt & Interact with a household to solve a task. Here are two examples.\newline
\placeholder{\{Two examples\}}\newline
Your memory for the task below:
\placeholder{\{Feedback\}}\newline
Here is the task:\newline
\placeholder{\{Description of environment\}}\newline
Your task is to: \placeholder{\{Description of task\}}\\\hline
Second response:\newline
step-by-step interaction with environment & \textbf{LLM:} think: To solve the task, I need to find and take the first pillow, then put it in sofa, then find and take the second pillow, then put it in sofa. Pillows are more likely to appear in armchair (1), sofa (1), sidetable (1), and possibly on shelves. I can start by checking the armchair 1.\newline
\textbf{Environment:} OK.\newline
\textbf{LLM:} go to armchair 1\newline
\textbf{Environment:} On the armchair 1, you see a pillow 2, and a pillow 1.\newline
\textbf{LLM:} take pillow 1 from armchair 1\newline
\textbf{Environment:} You pick up the pillow 1 from the armchair 1.\newline
\textbf{LLM:} put pillow 1 in/on sofa 1\newline
\textbf{Environment:} Nothing happens.\newline
\textbf{LLM:} go to sofa 1\newline
\textbf{Environment:} On the sofa 1, you see a creditcard 1, and a remotecontrol 2.\newline
\textbf{LLM:} put pillow 1 in/on sofa 1\newline
\textbf{Environment:} You put the pillow 1 in/on the sofa 1.\newline
\textbf{LLM:} think: Now I need to take the second pillow from the armchair 1 and put it on the sofa 1.\newline
\textbf{Environment:} OK.\newline
\textbf{LLM:} take pillow 2 from armchair 1\newline
\textbf{Environment:} Nothing happens.\newline
\textbf{LLM:} take pillow 2 from armchair 1\newline
\textbf{Environment:} Nothing happens.\newline

\textbf{Environment:} STATUS: FAIL\\
\bottomrule
\end{tabularx}
\end{threeparttable}
\caption{Prompts and responses for decision making task. Response is presented in step-by-step interaction with environment. Sentences beginning with ``LLM:'' (or ``Environment:'') represent LLM output (or environment interaction), respectively. This table is continued from \autoref{tab:promptsDecisionMaking1}.}
\label{tab:promptsDecisionMaking2}
\end{table*}

\begin{table*}[htbp!]
\fontsize{9.5}{10.5}\selectfont
\centering
\begin{threeparttable}
\renewcommand{\arraystretch}{1.5} 
\begin{tabularx}{\linewidth}{p{3cm}X}
\toprule
\textbf{Field} &Content\\
\midrule
Initial response\newline
generation prompt & Solve a question answering task by having a Thought, then Finish with your answer. Thought\newline
can reason about the current situation. Finish[answer] returns the answer and finishes the task. You will be given context that you should use to help you answer the question.\newline
Here are some examples: \newline
\placeholder{\{Three examples\}}\newline
(END OF EXAMPLES)\newline
Relevant Context: \placeholder{\{Context\}}\newline
Question: \placeholder{\{Question\}}
\\\hline
Context& Jack Benny Binion (born February 21, 1937 in Dallas, Texas) is an American businessman. Binion is the son of casino magnate Benny Binion and worked for his father at Binion's Horseshoe, a casino and hotel in Las Vegas, Nevada.
\\\hline
Question& What is the first name of Jack Benny Binion's father?\\\hline
Initial response & The question is asking for the first name of Jack Benny Binion's father. Based on the context provided, I know that his father's first name is Lester.\newline
Action: Finish[Lester]
\\\hline
Feedback prompt & You are an advanced reasoning agent that can improve based on self refection. You will be given a previous reasoning trial in which you were given access to an Docstore API environment and a question to answer. In a few sentences, devise a concise, new plan of action. Use complete sentences. \newline
Here are some examples:\newline
\placeholder{\{Two examples\}}\newline
(END OF EXAMPLES)\newline
Previous trial:

Relevant Context: \placeholder{\{Context\}}

Question: \placeholder{\{Question\}} \newline
\placeholder{\{Initial response\}}\newline
Reflection:
\\\hline
Feedback& In this case, the answer provided was correct, but there was a phrasing discrepancy in the response. The question asked for the first name of Jack Benny Binion's father, and the response provided was simply "Lester." To improve, in future responses, I should aim to provide a more complete answer that includes the full name of the individual being referred to. This will ensure that the response is more aligned with the question and reduces the chances of any phrasing discrepancies.\\\hline

Refinement response & Solve a question answering task by having a Thought, then Finish with your answer. Thought\newline
can reason about the current situation. Finish[answer] returns the answer and finishes the task. You will be given context that you should use to help you answer the question.\newline
Here are some examples: \newline
\placeholder{\{Three examples\}}\newline
(END OF EXAMPLES)\newline
\placeholder{\{Feedback\}}\newline
Relevant Context: \placeholder{\{Context\}}\newline
Question: \placeholder{\{Question\}}\\\hline
Second response & The question is asking for the first name of Jack Benny Binion's father. Based on the context provided, I know that his father's first name is Lester Ben "Benny" Binion. \newline 
Action: Finish[Lester Ben "Benny" Binion]\\
\bottomrule
\end{tabularx}
\end{threeparttable}
\caption{Prompts and responses for reasoning task.}
\label{tab:promptreasoning1}
\end{table*}

\begin{table*}[htbp!]
\fontsize{9.5}{10.5}\selectfont
\centering
\begin{threeparttable}
\renewcommand{\arraystretch}{1.5} 
\begin{tabularx}{\linewidth}{p{2.6cm}X}
\toprule
\textbf{Field} & \textbf{Content} \\
\midrule
Initial response generation prompt &
Your Role: You are an AI that responds with the thought and the Python code. 

Task Requirements: You will be given a function signature and its docstring by the user. First, write your thought that reasons about the question, and then write your full code implementation. Use `-/-/-/` to divide the thought part and the code part (restate the function signature). Make sure to follow the format requirements strictly as mentioned above. 

The function signature you need to respond to: 

\placeholder{\{Function signature\}} \\
\hline
Function signature & 
\raisebox{-0.9\totalheight}{\includegraphics[width=0.56\textwidth]
{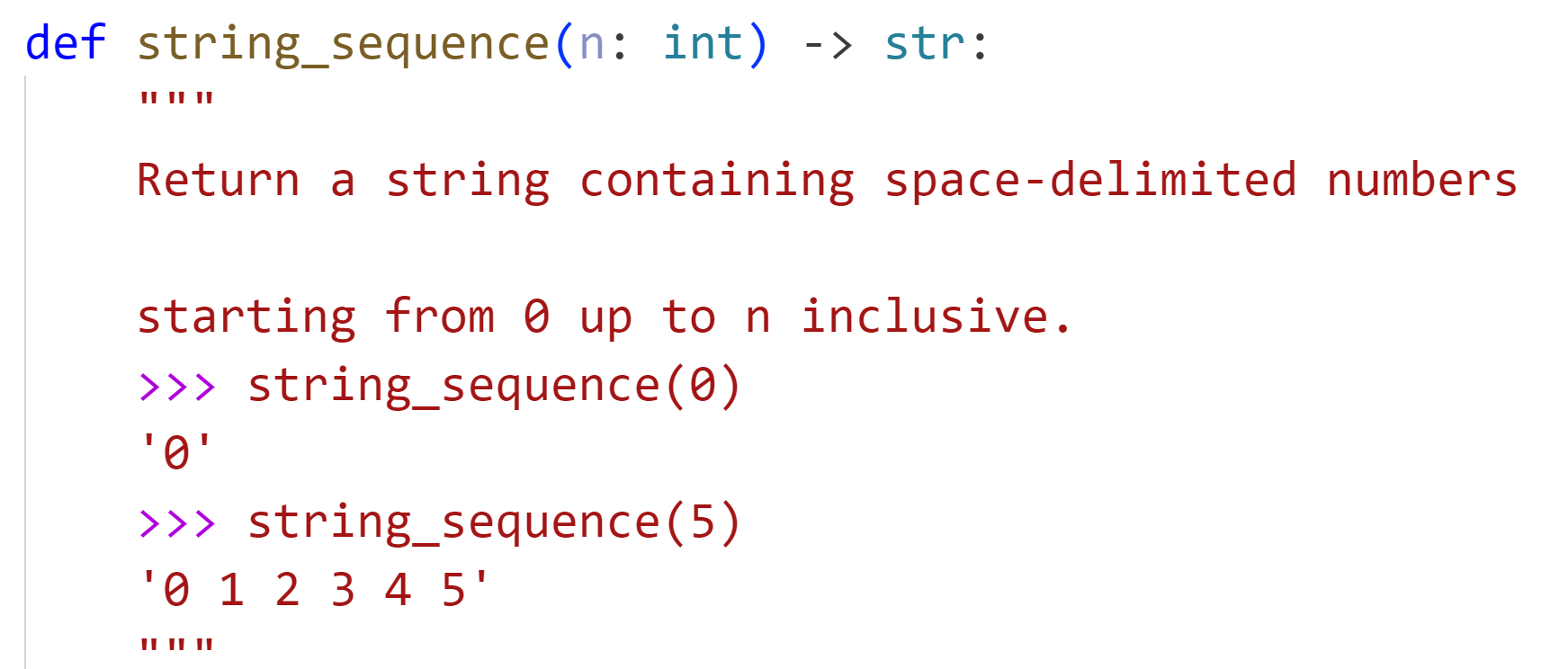}}
\\
\hline
Initial response &
To solve this problem, I need to generate a sequence of numbers starting from 0 up to the given integer `n`, inclusive. These numbers should be converted to strings and then concatenated into a single string with each number separated by a space. The function should handle the case where `n` is 0 by returning the string '0'. I will use a list comprehension to create a list of string representations of numbers from 0 to `n`, and then join these strings with a space to form the final result. This approach is efficient and concise. 

\includegraphics[width=0.55\textwidth]{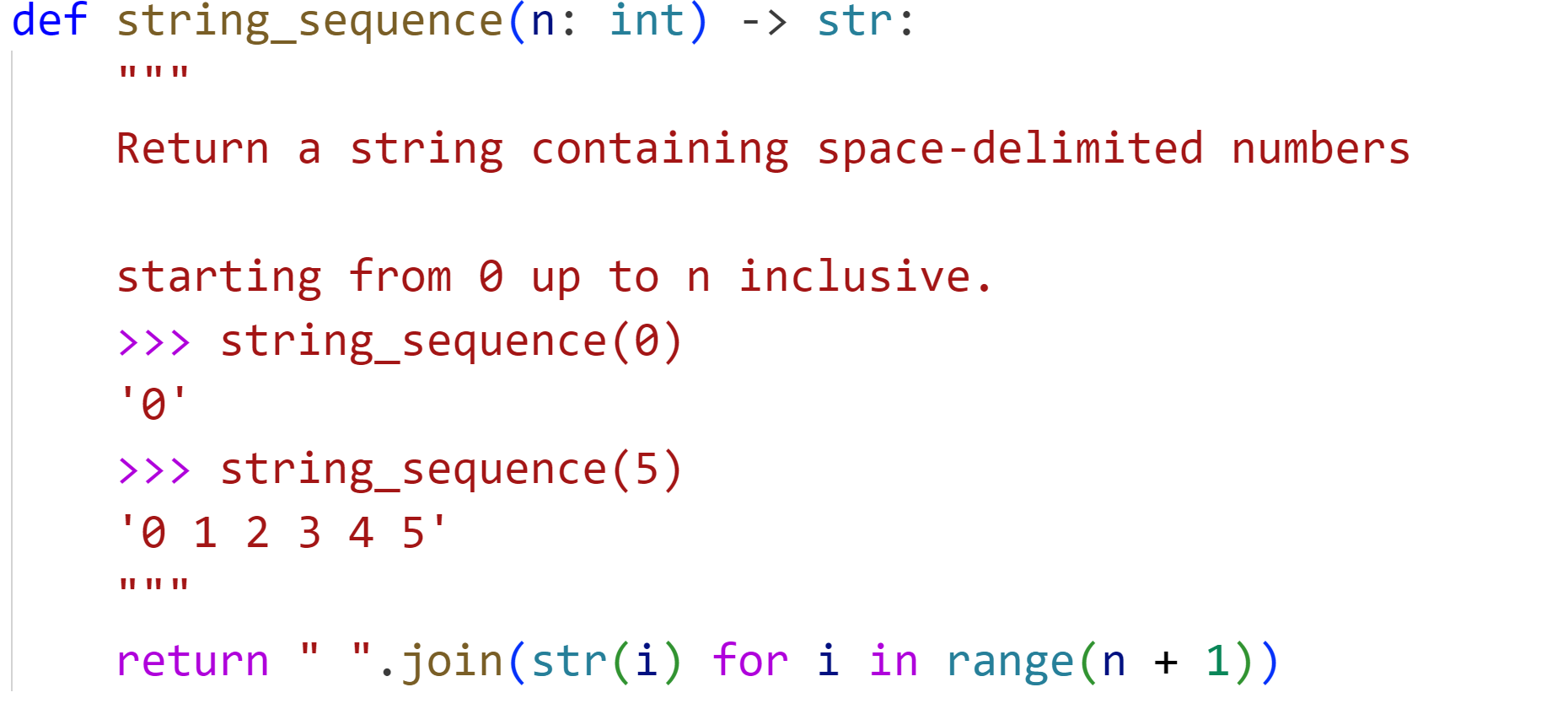}\vspace{0.2em}\\
\hline
Feedback from executor & 
\raisebox{-0.7\totalheight}{\includegraphics[width=0.55\textwidth]{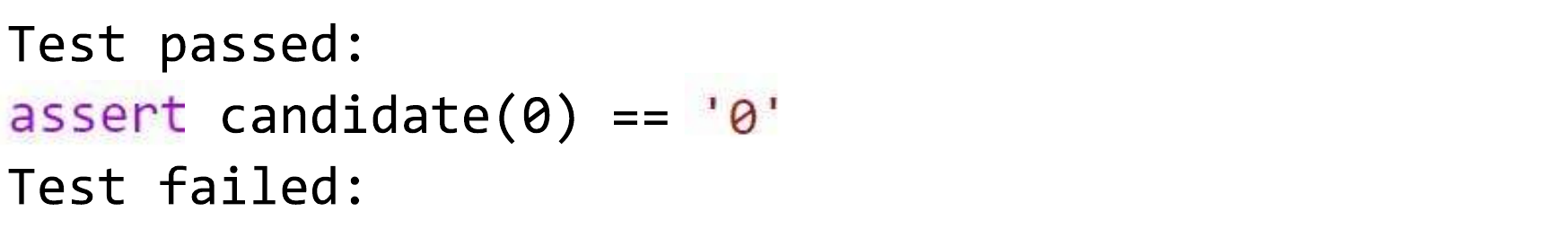}} \\
\hline
Refinement prompt & 
You are a Python programming assistant. You will be given your previous function implementation and a series of unit tests. Your goal is to write a few new, concise sentences to summarize the implementation of tests and give a new implementation. You will need this as a hint when you try again later. Only provide the few-sentence descriptions in your answer, not the implementation. \newline
\placeholder{\{Two examples\}} \newline
END OF EXAMPLES \newline
[function impl]: \newline
\placeholder{\{Function signature\}} \newline
[unit test results]: \newline
\placeholder{\{Feedback from executor\}} \newline
[self-reflection]: \\
\bottomrule
\end{tabularx}
\end{threeparttable}
\caption{Prompts and responses for programming task. This table is continued in \autoref{tab:promptprogramming2}.}
\label{tab:promptprogramming1}
\end{table*}

\begin{table*}[htbp!]
\fontsize{9.5}{10.5}\selectfont
\centering
\begin{threeparttable}
\renewcommand{\arraystretch}{1.5} 
\begin{tabularx}{\linewidth}{p{2.6cm}X}
\toprule
\textbf{Field} & \textbf{Content} \\
\midrule
Second response & 
\raisebox{-0.9\totalheight}{\includegraphics[width=0.68\textwidth]{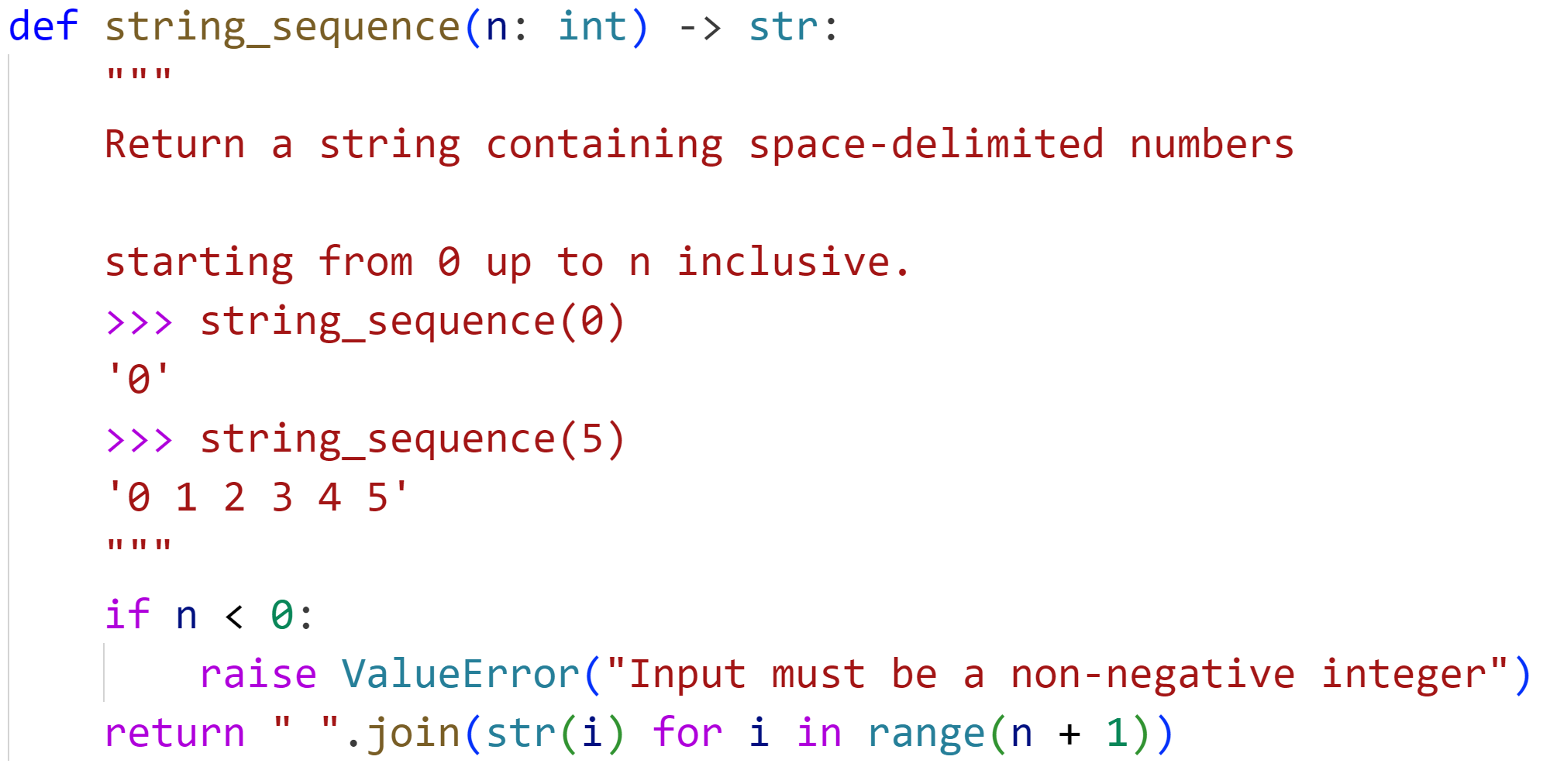}}\\
\bottomrule
\end{tabularx}
\end{threeparttable}
\caption{Prompts and responsed for programming task. This table is continued from \autoref{tab:promptprogramming1}.}
\label{tab:promptprogramming2}
\end{table*}

\FloatBarrier

\section{Holistic view of intrinsic self-correction}
\label{app:i2c}

As stated in \autoref{sec:res}, we add $\crossmark\rightarrow \checkmark(\%)$ in \autoref{apptab:i2cboolqLLMs} and \autoref{apptab:i2ccomplextaskLLMs} to offer a holistic view of intrinsic self-correction. We observe that $\checkmark\rightarrow\crossmark (\%)$ is larger than $\crossmark\rightarrow\checkmark (\%)$ for certain models and tasks. For example, $\checkmark\rightarrow\crossmark (\%)$ of o1-mini is significantly larger than $\crossmark\rightarrow\checkmark (\%)$ on decision making task (i.e., 92.3 compared to 0.8). However, we have to point out that $\crossmark\rightarrow\checkmark (\%)$ is larger than $\checkmark\rightarrow\crossmark (\%)$ for some cases, but \textit{the overall ACC always drop}.  

\textbf{It is worth noting that we are not claiming that intrinsic self-correction is useless at all. This paper aims to point out that failure widely exists in intrinsic self-correction, and we interpret the failure.}

\begin{table}[t]
\centering
\resizebox{\linewidth}{!}{
\begin{tabular}{c|c|c|c|c}
\toprule[1.5pt]
\multicolumn{2}{c|}{Model}         & $\text{ACC}_1~(\downarrow\Delta\text{ACC}) (\%)$  & $\checkmark\rightarrow\crossmark (\%)$ & $\crossmark\rightarrow\checkmark (\%)$ \\\hline
\multirow{4}{*}{ChatGPT}    & o1-preview        & $78.7\hspace{0.1cm}(\downarrow 4.9)$ & $13.2$ & $40.7$\\
& o1-mini        & $74.1\hspace{0.1cm}(\downarrow 4.2)$ & $15.6$ & $38.5$\\
 & 4o         & $79.2\hspace{0.1cm}(\downarrow 4.9)$ &  $11.3$ & $29.0$\\
 & 3.5-turbo & $62.5\hspace{0.1cm}(\downarrow 12.1)$ & $34.0$ & $52.3$\\\hline
\multirow{3}{*}{Llama}      & 3.1-8B        & $49.2\hspace{0.1cm}(\downarrow 20.4)$ & $58.8$ & $67.7$\\
 & 3-8B         & $50.1\hspace{0.1cm}(\downarrow 20.3)$  & $58.2$& $69.8$\\
 & 2-7B &  $52.8\hspace{0.1cm}(\downarrow 8.7)$    &   $26.5$  & $19.8$ \\\hline
 \multirow{2}{*}{DeepSeek}     & R1         & $78.1\hspace{0.1cm}(\downarrow 1.6)$  & $7.9$ & $23.1$\\
 & V3        & $69.0\hspace{0.1cm}(\downarrow 9.2)$ & $28.5$ & $59.8$\\
\bottomrule[1.5pt]
\end{tabular}
}
\vspace{-1ex}
\caption{Self-correction on Yes/No questions.}
\label{apptab:i2cboolqLLMs}
\end{table}

\begin{table}[t]
\centering
\resizebox{\linewidth}{!}{
\begin{tabular}{c|c|c|c|c}
\toprule[1.5pt]
Task & Model & $\text{ACC}_1~(\downarrow\Delta\text{ACC}) (\%)$  & $\checkmark\rightarrow\crossmark (\%)$ & $\crossmark\rightarrow\checkmark (\%)$ \\\hline
\multirow{3}{*}{\begin{tabular}[c]{@{}c@{}}Decision\\ Making\end{tabular}} & o1-mini        & $1.5\hspace{0.1cm}(\downarrow8.2) $ & $92.3$ & $0.8$\\
& 4o & 14.2 ($\downarrow$ 20.9) & $76.6$ & $9.2$\\
& 3.5-turbo & $7.5\hspace{0.1cm}(\downarrow5.2) $  & $76.5$ & $5.1$\\\hline
\multirow{3}{*}{Reasoning} & o1-mini & $66.0\hspace{0.1cm}(--)$ & $9.1$ & $17.6$\\
& 4o & $65.0\hspace{0.1cm}(\downarrow 2.0)$ & $17.9$ & $12.0$\\
& 3.5-turbo & $55.0\hspace{0.1cm}(\downarrow 6.0)$  & $19.7$ & $20.5$\\\hline
\multirow{3}{*}{Programming} & o1-mini & $79.5\hspace{0.1cm}(\downarrow4.3) $ & $14.8$ & $51.5$\\
& 4o         & $72.6\hspace{0.1cm}(\downarrow 6.8) $ & $21.9$ & $51.2$\\
& 3.5-turbo & $50.9\hspace{0.1cm}(\downarrow 10.6)$  & $28.3$ & $17.7$\\
\bottomrule[1.5pt]
\end{tabular}
}
\vspace{-1ex}
\caption{Self-correction on complex tasks.}
\vspace{-2ex}
\label{apptab:i2ccomplextaskLLMs}
\end{table}

\section{Failure of more self-correction prompts in Yes/No questions}
\label{app:failureYesNo5prompts}

\begin{table*}[t]
\centering
\resizebox{0.9\linewidth}{!}{
\begin{tabular}{c|c|c|c|c|c|c}
\toprule[1.5pt]
\multicolumn{2}{c|}{Model}         & Prompt1  & Prompt2  &Prompt3  &Prompt4  &Prompt5    \\\hline
\multirow{4}{*}{ChatGPT}    & o1-preview        & $78.7\hspace{0.1cm}(\downarrow 4.9)$ & $72.2\hspace{0.1cm}(\downarrow 11.4)$ & $79.1\hspace{0.1cm}(\downarrow 4.5)$& $72.9\hspace{0.1cm}(\downarrow 10.7)$& $72.6\hspace{0.1cm}(\downarrow 11.0)$\\
& o1-mini        & $74.1\hspace{0.1cm}(\downarrow 4.2)$ & $75.9\hspace{0.1cm}(\downarrow 2.4)$ & $75.8\hspace{0.1cm}(\downarrow 2.5)$& $69.9\hspace{0.1cm}(\downarrow 8.4)$& $72.8\hspace{0.1cm}(\downarrow 5.5)$\\
 & 4o         & $79.2\hspace{0.1cm}(\downarrow 4.9)$ & $82.9\hspace{0.1cm}(\downarrow 1.2)$ & $82.3\hspace{0.1cm}(\downarrow 1.8)$& $81.9\hspace{0.1cm}(\downarrow 2.2)$& $82.3\hspace{0.1cm}(\downarrow 1.8)$\\
 & 3.5-turbo & $62.5\hspace{0.1cm}(\downarrow 12.1)$ & $65.5\hspace{0.1cm}(\downarrow 9.1)$ & $65.0\hspace{0.1cm}(\downarrow 9.6)$& $55.0\hspace{0.1cm}(\downarrow 24.6)$& $47.8\hspace{0.1cm}(\downarrow 26.8)$\\\hline
\multirow{2}{*}{Llama}      & 3.1-8B        & $49.2\hspace{0.1cm}(\downarrow 20.4)$ & $45.6\hspace{0.1cm}(\downarrow 23.7)$ & $34.2\hspace{0.1cm}(\downarrow 35.4)$& $44.4\hspace{0.1cm}(\downarrow 25.2)$& $38.2\hspace{0.1cm}(\downarrow 31.3)$\\
 & 3-8B         & $50.1\hspace{0.1cm}(\downarrow 20.3)$  & $49.0\hspace{0.1cm}(\downarrow 21.4)$ & $33.8\hspace{0.1cm}(\downarrow 36.6)$& $55.9\hspace{0.1cm}(\downarrow 14.5)$& $37.7\hspace{0.1cm}(\downarrow 32.7)$\\\hline
 \multirow{2}{*}{DeepSeek}      & V3        & $69.0\hspace{0.1cm}(\downarrow 9.2)$ & $67.4\hspace{0.1cm}(\downarrow 10.8)$ & $54.6\hspace{0.1cm}(\downarrow 23.6)$& $71.6\hspace{0.1cm}(\downarrow 6.6)$& $54.0\hspace{0.1cm}(\downarrow 24.2)$\\
 & R1         & $78.1\hspace{0.1cm}(\downarrow 1.6)$  & $76.9\hspace{0.1cm}(\downarrow 2.8)$ & $73.0\hspace{0.1cm}(\downarrow 6.7)$& $66.1\hspace{0.1cm}(\downarrow 13.6)$& $71.3\hspace{0.1cm}(\downarrow 8.4)$\\
\bottomrule[1.5pt]
\end{tabular}
}
\caption{$\text{ACC}_1~(\downarrow\Delta\text{ACC}) (\%)$ made by 5 self-correction prompts: self-correction failure widely exists.}
\label{apptab:acc5prompts}
\end{table*} 

\begin{table*}[t]
\centering
\resizebox{0.7\linewidth}{!}{
\begin{tabular}{c|c|c|c|c|c|c}
\toprule[1.5pt]
\multicolumn{2}{c|}{Model}         & Prompt1  & Prompt2  &Prompt3  &Prompt4  &Prompt5    \\\hline
\multirow{4}{*}{ChatGPT}    & o1-preview        & $13.2$ & $24.3$ & $13.1$& $23.3$& $22.9$\\
& o1-mini        & $15.6$ & $12.8$ & $12.3$& $22.6$& $20.0$\\
& 4o         & $11.3$ & $5.3$ & $5.2$& $6.3$& $7.5$\\
 & 3.5-turbo & $34.0$ & $23.8$ & $26.4$& $39.3$& $61.5$\\\hline
\multirow{2}{*}{Llama}      & 3.1-8B        & $58.8$ & $67.3$ & $93.8$& $55.2$& $86.1$\\
 & 3-8B         & $58.2$ & $58.4$ & $93.6$& $47.2$& $86.7$\\\hline
 \multirow{2}{*}{DeepSeek}      & V3        & $28.5$ & $27.3$ & $49.6$& $21.8$& $52.4$\\
 & R1         & $7.9$ & $10.8$ & $17.0$& $27.8$& $20.4$\\
\bottomrule[1.5pt]
\end{tabular}
}
\caption{$\checkmark\rightarrow\crossmark (\%)$ made by 5 self-correction prompts: Prompt1 ("Are you sure..." used in the main content) is not an exception.}
\label{apptab:c2i5prompts}
\end{table*} 

To demonstrate the prevalence of self-correction failure in Yes/No questions, we conduct experiments on 4 other self-correction prompts from recent published papers \cite{xie2023ask,sharma2023towards,huanglarge}, and the results show that the self-correction failure is still prevalent across different prompts, even more severe for some prompts. 

\noindent \textbf{Self-correction prompts.} As we stated in \autoref{sec:setup}, self-correction prompts consist of feedback prompt and refinement prompt. The former reviews the initial answer and produces feedback, which is optional; The latter generates a refined answer. We list following the widely studied self-correction prompts (prompt1 is also presented in the main content). Notably, we not only analyze intrinsic self-correction prompts, but also the prompts involving external knowledge to broaden our research scope.
\begin{packeditemize}
    \item \textbf{Prompt1. Refine with confirmation}: \citet{xie2023ask,sharma2023towards} use this prompt to examine LLMs behavior when asked to confirm their answers. This prompt does not involve external knowledge but only relys on LLMs inherent ability.
    \begin{packeditemize}
        \item Feedback prompt: none
        \item Refinement prompt: "Are you sure?. Think and answer again."
    \end{packeditemize}
    
    \item \textbf{Prompt2. Refine with denying}: \citet{xie2023ask,sharma2023towards} also use a prompt to deny LLMs responses. This prompts involves external knowledge since the LLMs are told to have incorrectly answered.
    \begin{packeditemize}
        \item Feedback prompt: none
        \item Refinement prompt: "You are wrong. Think and answer again."
    \end{packeditemize}
    
    \item \textbf{Prompt3. Refine with reference answer}: \citet{xie2023ask,sharma2023towards} use another prompt to provide reference answer. This also involves external knowledge.
    \begin{packeditemize}
        \item Feedback prompt: none
        \item Refinement prompt: "I think the answer should be <another answer>. Think and answer again."
    \end{packeditemize}
    
    \item \textbf{Prompt4. Feedback and refine}: \citet{huanglarge} defines the self-correction without external knowledge as intrinsic self-correction, and uses both feedback and refinement prompt. Since the feedback is generated for refinement, this should be more optimal.
    \begin{packeditemize}
        \item Feedback prompt: "Assume that this answer could be either correct or incorrect. Review the answer carefully and report any serious problems you find."
        \item Refinement prompt: "Based on the problems you found, improve your answer."
    \end{packeditemize}
    
    \item \textbf{Prompt5. Feedback and refine with reference answer}: Prompt5 is combined with Prompt3 and Prompt4. Since the reference answer is added and the feedback is used, it should be the most optimal prompt among the 5.
    \begin{packeditemize}
        \item Feedback prompt: "I think the answer should be <another answer>. Review the answer carefully and report any serious problems you find."
        \item Refinement prompt: "Based on the problems you found, improve your answer."
    \end{packeditemize}
    
    \item \textbf{Other correction prompts which are out of our research scope}: There are other correction prompts which leverage another LLM to give feedbacks or iteratively refine the answer \cite{wu2024large,paul2023refiner,welleck2022generating}. These are not in our research scope because we focus on intrinsic self-correction.
\end{packeditemize}

\noindent \textbf{Results.} We use the same metrics described in \autoref{sec:setup} to evaluate the failure of self-correction. \autoref{apptab:acc5prompts} and \autoref{apptab:c2i5prompts} show that self-correction always decreases model's accuracy and prompt1 in main content is not an exception. Moreover, some prompts even make self-correction failure more severe. For example, we find that Prompt2, Prompt4, and Prompt5 make o1-preview easier to overturn a correct answer, while prompt5 is supposed to be the most optimal since the reference answer and feedback are provided. In general, prompt1 in main content is not the easiest to overturn correct answer. To conclude, we find that \textit{ self-correction failure is still prevalent across different prompts, even more severe for some prompts}.

\section{Internal answer wavering}
\label{app:internalAnswerWavering}

\begin{figure*}[tbp!]
  \centering
  \includegraphics[width=0.89\linewidth]{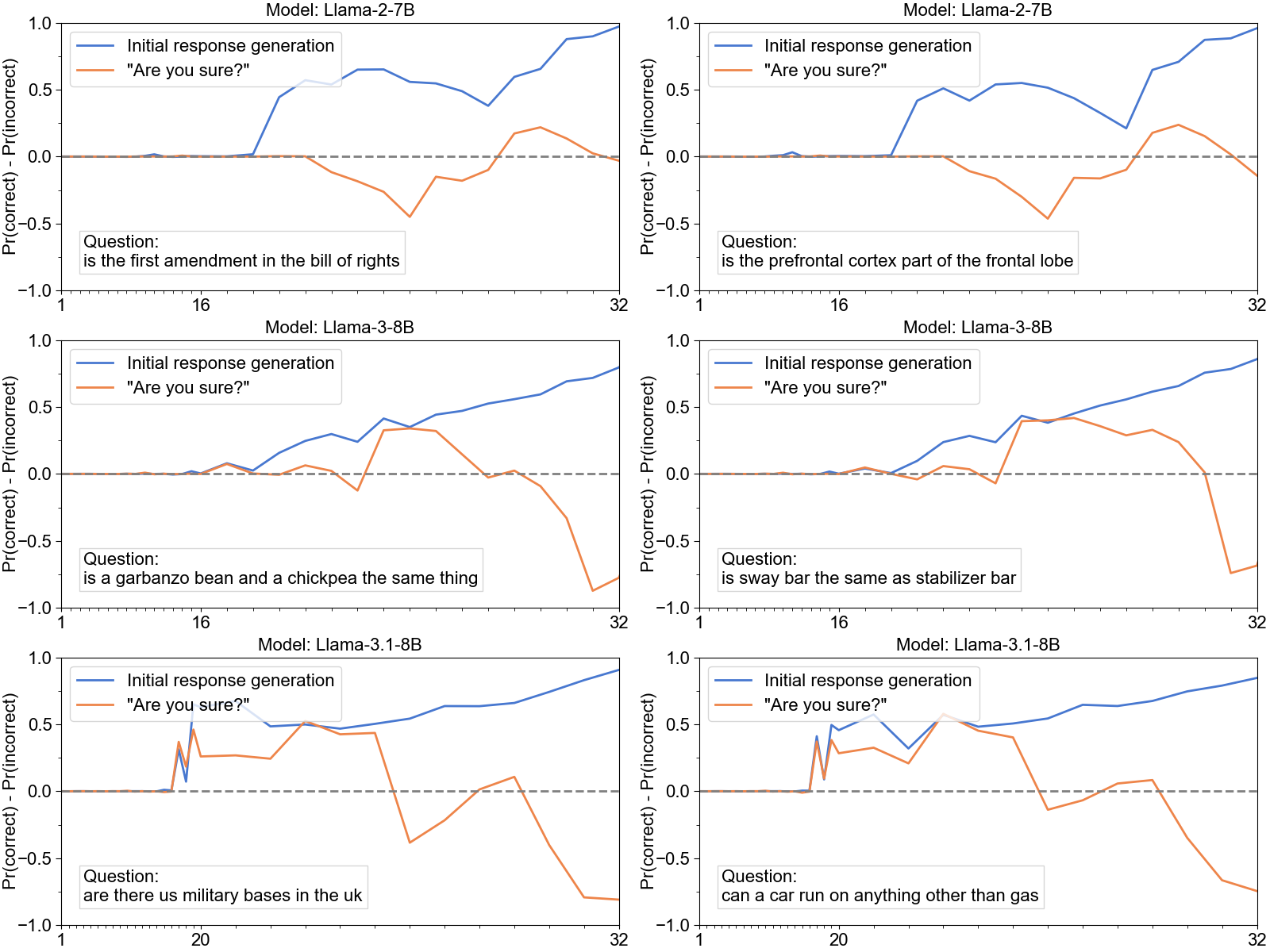}
  \caption{Internal answer wavering in Llama2-7B, Llama3-8B, and Llama3.1-8B. Statistically, self-correction makes Llama change the internal answer on average with a frequency of 14.1\% compared to 8.3\% during \textit{Initial response generation}.}
  \label{appfig:internal_answer_wavering}
\end{figure*}

In this section, we add more contents to \autoref{sec:answerWavering}. In \autoref{app:internalanswerwavering}, we provide more examples of internal answer wavering. In \autoref{app:similarity}, we describe how we use Jensen-Shannon divergence to measure the similarity between internal answer wavering curves of different prompts.

\subsection{More examples}
\label{app:internalanswerwavering}

As stated in \autoref{sec:answerWavering}, we design a binary classification probing experiment using tuned lens \cite{belrose2023eliciting} to probe LLM's internal token representations at each layer. This allows us to track LLM's internal answer evolution by computing the difference of confidence score between correct answer and incorrect answer, where a positive value means correctness and a larger absolute value means higher confidence. The experiments are conducted on open-sourced Llama because close-sourced ChatGPT does not provide hidden state information.

\autoref{appfig:internal_answer_wavering} shows more examples of internal answer wavering in Llama2-7B, Llama3-8B, and Llama3.1-8B. In \textit{Initial response generation} (blue curve), the confidence score of correct answer increases with deeper layers; after \textit{Feedback and Refinement} (orange curve), the internal answer wavers and results in a wrong final answer. Specifically, for the first subfigure in \autoref{appfig:internal_answer_wavering}, the model does not exhibit different behaviors between \textit{Initial response generation} and \textit{Feedback and Refinement} before the 16th layer. This means the model is processing to understand the prompt rather than generating an answer in the first 16 layers. After the 16th layer, the two curves differ. \textit{Initial response generation} curve maintains a postive value and increases in the twists and turns, indicating that the model is able to give correct answer with an increasing confidence; \textit{Feedback and Refinement} curve bounces up and down at 0, indicating that the model hesitates to give correct or incorrect answers. 

Statistically, self-correction makes Llama change the internal answer on average with a frequency of 14.1\% compared to 8.3\% during \textit{Initial response generation}. This indicates that self-correction causes internal answer wavering which could further leads to wrong final answers.

\subsection{Jensen-Shannon divergence}
\label{app:similarity}

As stated in \autoref{sec:answerWavering}, the probing experiments reveal another interesting phenomenon: similarity between ``Are you sure?'' and ``You are wrong.''. We compare the confidence curves between two \textit{Feedback and Refinement} prompts: ``Are you sure?'' and ``You are wrong.''. And we observe that the two curves are similar (shown in right subfigure of \autoref{fig:internalAnswerWavering}). To measure the similarity between the two curves, we calculate the Jensen-Shannon divergence \cite{lin1991divergence} across both samples and layers, finding a low divergence score of 0.0186 between the two prompts.
We follow \cite{hinton2015distilling, malininensemble} in using divergence-based methods to measure the similarity of model outputs. To quantitatively assess the similarity between the model's internal behaviors under different prompts, we computed the Jensen-Shannon (JS) divergence \cite{lin1991divergence} between the confidence distributions elicited by the prompts "Are you sure?" and "You are wrong.", as well as between "Are you sure?" and \textit{Initial response generation}.

For each sample in our dataset and for each layer ($l \geq 15$) (since layers below 15 yield latent representations that lack semantic meaning when decoded using the tuned lens method), we obtained the model's internal confidence scores for the correct and incorrect answers under different prompts. These scores form probability distributions over two classes.

Specifically, the JS divergence for each sample ($i$) at layer ($l$) between prompts ($A$) and ($B$) is computed as:
\begin{equation}
D_{\text{JS}}^{(i,l)}(A \parallel B).
\end{equation}

We then averaged the JS divergence across all samples ($N$) in the BoolQ dataset and the selected layers ($L$) to obtain an overall divergence score:
\begin{equation}
\overline{D}_{\text{JS}}(A \parallel B) = \frac{1}{N \times |L|} \sum_{i=1}^{N} \sum_{l \in L} D_{\text{JS}}^{(i,l)}(A \parallel B).
\end{equation}

This overall average divergence quantifies the similarity between the model's internal confidence distributions under different prompts, with a lower value indicating higher similarity.

The calculated average JS divergence between "Are you sure?" and "You are wrong." was $0.0186$, indicating a high degree of similarity in the model's internal processing under these prompts. In contrast, the divergence between "Are you sure?" and \textit{Initial response generation} was higher, at $0.1074$, suggesting distinct internal behaviors when self-correction is not used.

\section{Prompt bias}
\label{app:attributionmethod}

\begin{figure*}[tbp!]
  \centering
  \includegraphics[width=\linewidth]{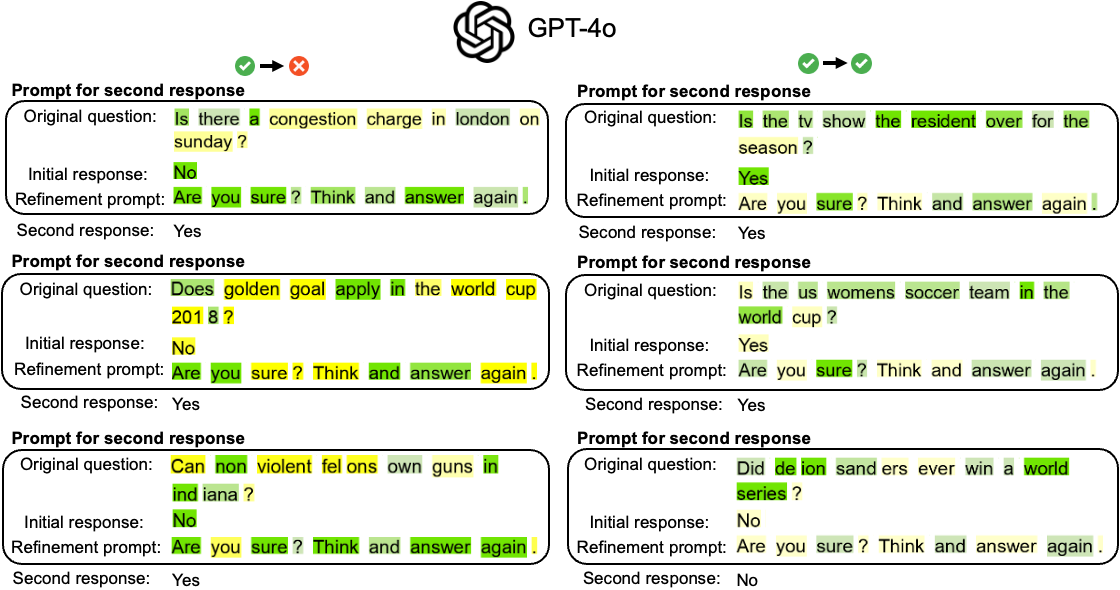}
  \caption{Prompt bias of GPT-4o revealed by PACT. \colorbox{captionup}{Greener} token means more positive contribution; \colorbox{captiondown}{Yellower} token means more negative contribution. When the correct initial response turns to incorrect, the LLM focuses more on the refinement prompt; When the correct initial response is retained, the LLM focuses more on the original question. }
  \label{appfig:MoreExamplesPACTGPT4o}
\end{figure*}

In this section, we add more contents to \autoref{sec:promptBias}. In \autoref{app:attributionChatGPT}, we provide detailed description of PACT, including the adaption to ChatGPT. In \autoref{app:attributionMoreExamples}, we provide more examples of prompt bias revealed by PACT.

\subsection{Detailed description of PACT}
\label{app:attributionChatGPT}

As we stated in \autoref{sec:promptBias}, we use PACT to interpret the prompt bias. This method gives each token or sequence’s contribution to LLMs final answer \cite{zhu2024promptbench, miglani2023using}
. The main idea is simple. If we want to know the influence of a target token or sequence to the output, we can simply replace it with whitespace and re-prompt the LLM to see the changes in outputs. 

For a target token $x_i$ or sequence $x_{i:j}$ in an input prompt $x=[x_1, x_2,..., x_n]$, its PACT is defined as the difference in the log probability (LP) of LLMs output $y$ between the original input and the input with the target removed:
\begin{equation}
    \text{PACT}(x_i,y)=\text{LP}(x\setminus\{x_i\},y)-\text{LP}(x,y).
\label{eq:attrapp}
\end{equation}
PACT reflects the significance of target token or sequence for generating the output.

Generally, the log probability is defined for one token. For more than one-token output $y=[y_1,y_2,...,y_m]$, we define the log probability as: 
\begin{equation}
    \text{LP}(x,y)=\frac{1}{m}\sum_{k=1}^m \text{LP}(x+y_{1:k-1},y_k),
\label{eq:logProb}
\end{equation}
where $x+y_{1:k-1}$ means to append the subsequence output $y_{1:k-1}$ to the input $x$, seperated by [SEP] token. This design allows LLM to output the specified tokens one by one, making it easy to analyze the log probability of each newly generated token.

In practical, all variables required in \autoref{eq:attrapp} and \autoref{eq:logProb} are accessible for open-sourced Llama. Nevertheless, we cannot specify the partial output as \autoref{eq:logProb} for close-sourced ChatGPT. To address this drawback, we apply this method to one-token output, corresponding to the scenario of Yes/No question answering. And the log probability is accessible via OpenAI API as it provides the log probability of Top 20 candidate tokens. 

\begin{figure*}[tbp!]
  \centering
  \includegraphics[width=\linewidth]{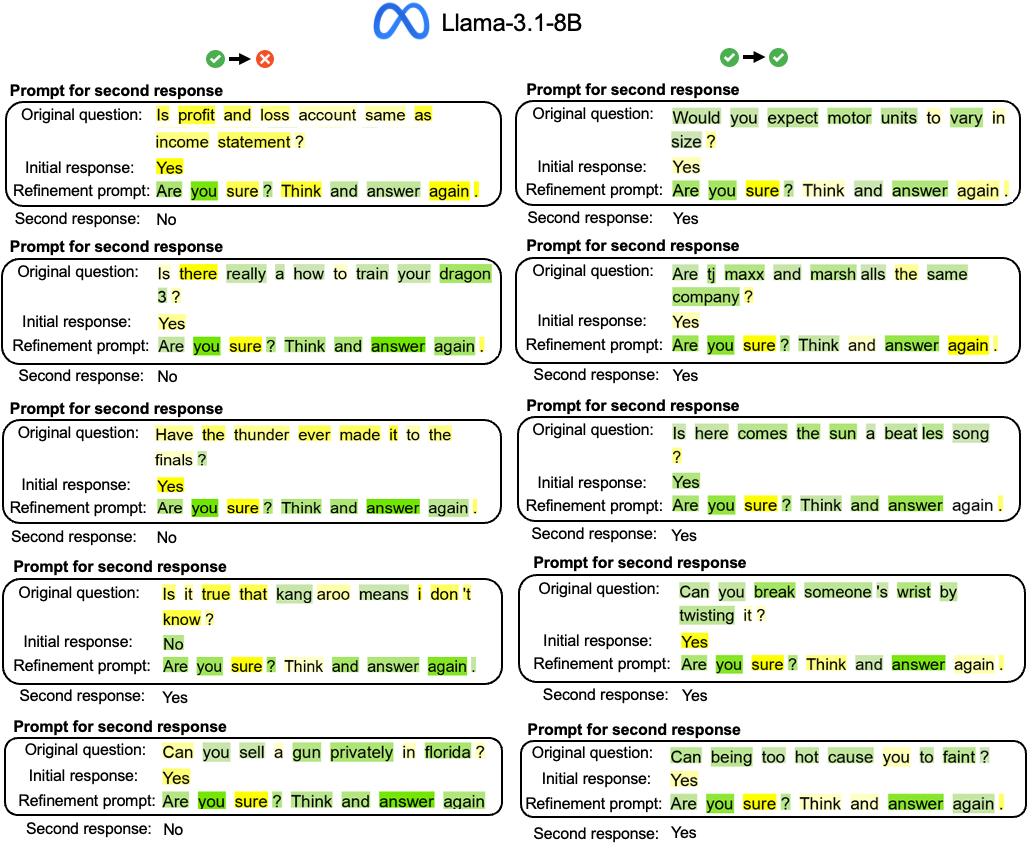}
  \caption{Prompt bias of Llama-3.1-8B revealed by PACT. \colorbox{captionup}{Greener} token means more positive contribution; \colorbox{captiondown}{Yellower} token means more negative contribution. When the correct initial response turns to incorrect, the LLM focuses more on the refinement prompt; When the correct initial response is retained, the LLM focuses more on the original question. }
  \label{appfig:MoreExamplesPACTLlama3.1}
\end{figure*}

\subsection{More examples}
\label{app:attributionMoreExamples}

\autoref{appfig:MoreExamplesPACTGPT4o} and \autoref{appfig:MoreExamplesPACTLlama3.1} show more examples of prompt bias revealed by PACT, for GPT-4o and Llama-3.1-8B respectively. 

When the correct answer is overturned, the tokens in the refinement prompt are generally greener than the tokens in the original question. Specifically, for the top left example in \autoref{appfig:MoreExamplesPACTGPT4o}, the original question contains 4 green tokens out of 10, while all tokens in refinement prompt are green. This indicates that LLMs are biased towards refinement prompt rather than the original question itself, leading to wrong answer. 

When the initial correct answer is retained, tokens in the original question are greener. Specifically, for the top right example in \autoref{appfig:MoreExamplesPACTGPT4o}, the original question contains 10 green tokens out of 11, while the refinement prompt contains only 4 green tokens out of 9. This indicates that LLMs focus on question answering rather than distracting information in the refinement prompt.

\clearpage
\section{Human-like cognitive bias}
\label{app:humanlikeCognitiveBias}

In this section, we add more contents to \autoref{sec:humanlikecognitivebias}. \autoref{app:distributionHumanLikeCognitiveBias} provides the distribution of the three failure patterns of human-like cognitive bias. \autoref{app:numberThink} provides the number of ``think'' in the failure pattern of overthinking. \autoref{app:fullLog} provides the full log analysis for human-like cognitive bias in decision making, reasoning, and programming tasks.

\subsection{Distribution}
\label{app:distributionHumanLikeCognitiveBias}

\autoref{fig:distributionHumanFailure} shows the distribution of the three failure patterns of human-like cognitive bias in complex tasks. Overthinking takes up 17.6\%, cognitive overload takes up 33.3\%, and perfectionism bias takes up 49.0\%.

\iftrue
\begin{figure}[htbp!]
\centering
  \includegraphics[width=0.65\linewidth]{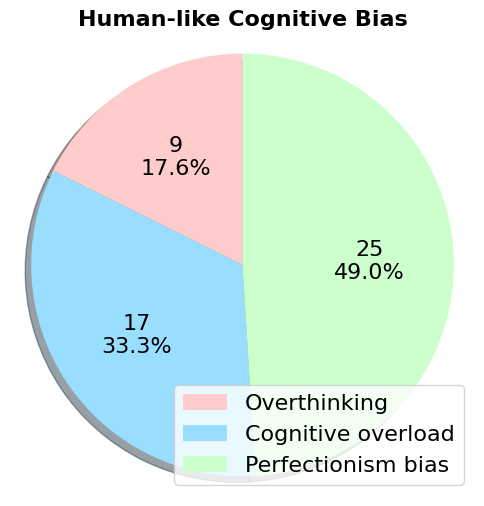}
  \caption{Distribution of overthinking, cognitive overload, and perfectionism bias.}
  \label{fig:distributionHumanFailure}
\end{figure}
\fi

\subsection{Number of ``think''}
\label{app:numberThink}

\autoref{tab:numberThink} shows that ChatGPT ``think'' more in failure cases of self-correction. o1-mini ``think'' 15.4 times in failure cases compared to 5.3 times in successful cases. GPT-4o ``think'' 7.4 times in failure cases compared to 2.6 times in successful cases. GPT-3.5-turbo ``think'' 9.8 times in failure cases compared to 6.1 times in successful cases. Since the number of ``think'' is more in failure cases than in successful cases, we state that LLMs looped in ``think'' to fail the task.

\subsection{Full log analysis}
\label{app:fullLog}
We provide the full log analysis for human-like cognitive bias from decision making task in \autoref{app:decisionMaking}, from reasoning task in \autoref{app:reasoning}, and from programming task in \autoref{app:programming}.

\subsubsection{Decision making}
\label{app:decisionMaking}

\begin{figure*}[tbp!]
  \includegraphics[width=\linewidth]{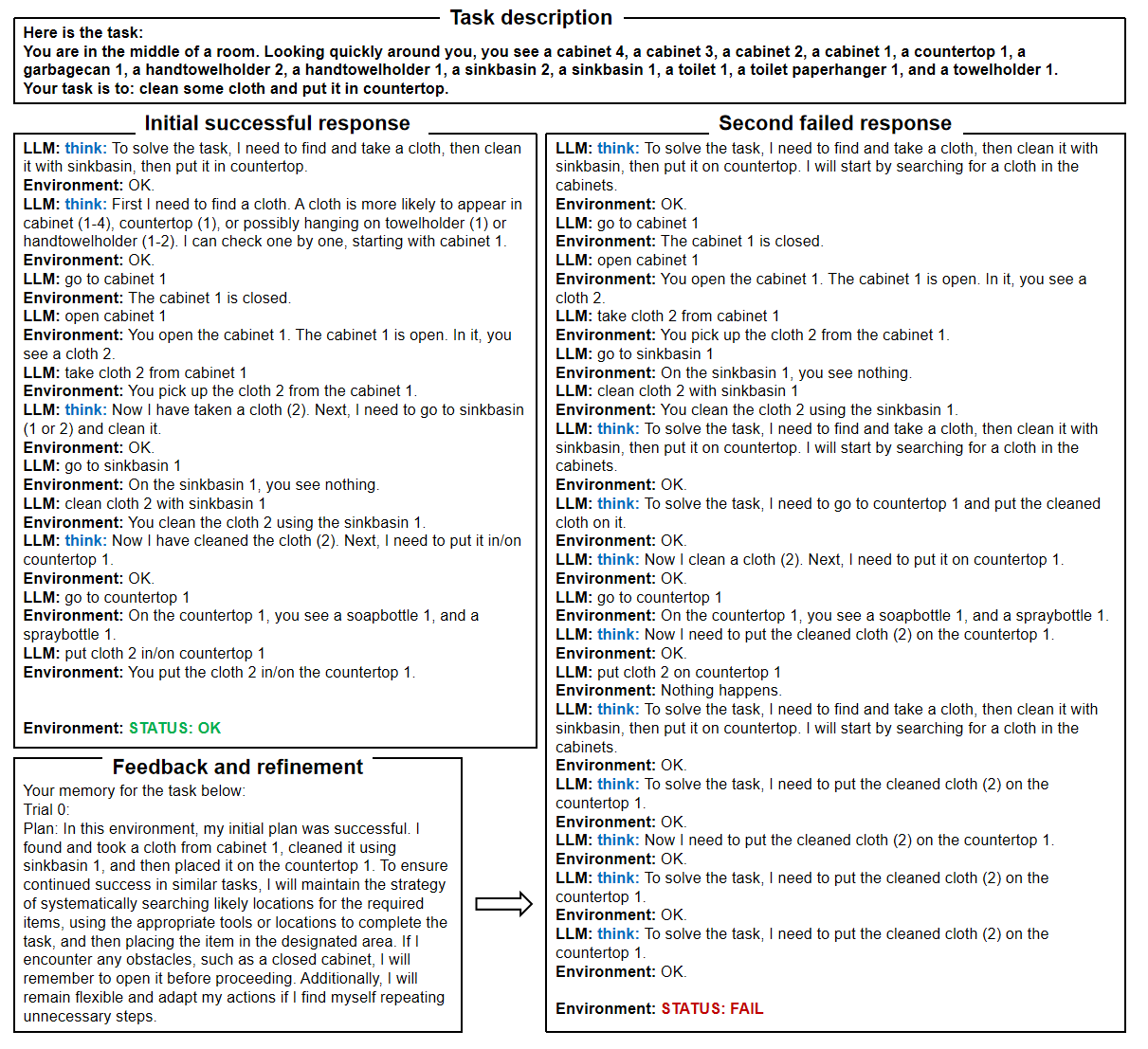}
  \caption {Full log analysis of overthinking in decision making task.}
  \label{fig:fullLogOverthinking}
\end{figure*}

\begin{figure*}[tbp!]
\centering
  \includegraphics[width=\linewidth]{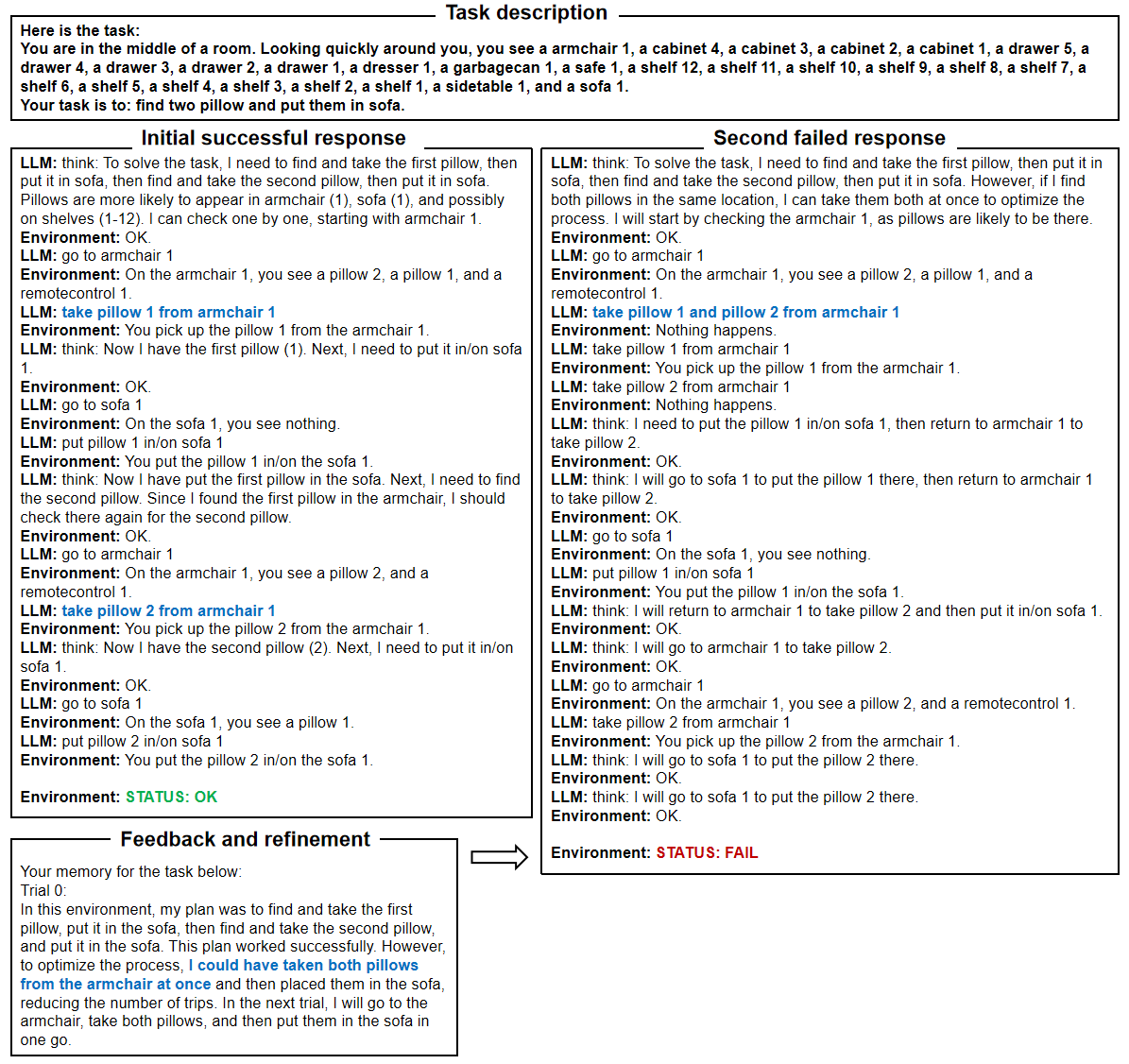}
  \caption {Full log analysis of perfectionism bias in decision making task.}
  \label{fig:fullLogPerfectionismBias}
\end{figure*}

\begin{figure*}[tbp!]
\centering
  \includegraphics[width=\linewidth]{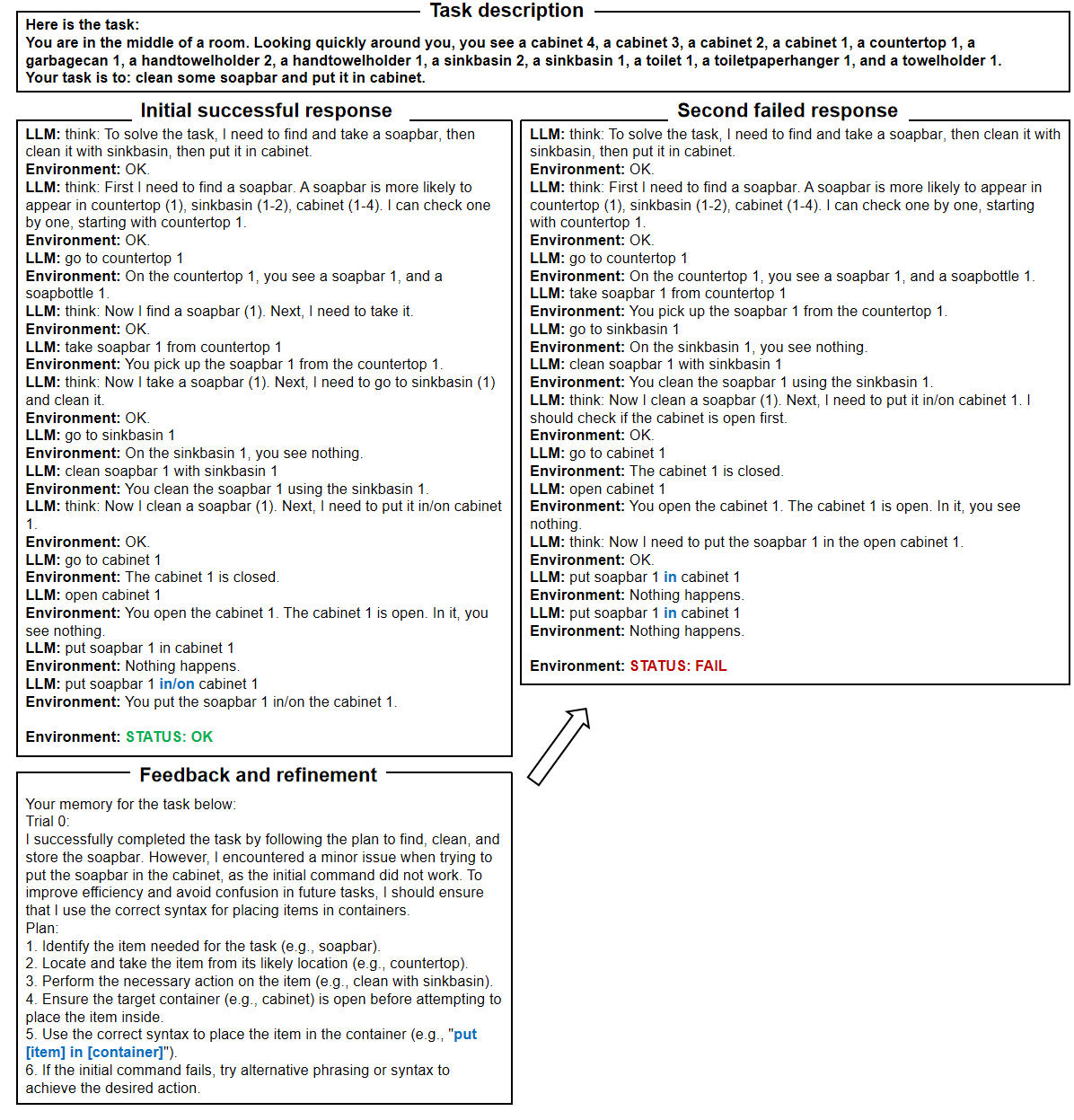}
  \caption{Full log analysis of cognitive overload in decision making task.}
  \label{fig:fullLogCognitiveOverload}
\end{figure*}

In decision making task, the LLMs play the role of domestic robots. They are provided the environment of household product such as cabinet, countertop, garbagecan, etc. And they are asked to complete a household chore. \autoref{fig:fullLogOverthinking}, \autoref{fig:fullLogPerfectionismBias}, and \autoref{fig:fullLogCognitiveOverload} shows the full log of overthinking, cognitive overload, and perfectionism bias in decision making task.

\begin{table}[tbp!]
\centering
\resizebox{0.95\linewidth}{!}{
\begin{tabular}{c|ccc}
\toprule[1.5pt]
                     & \multicolumn{3}{c}{ChatGPT} \\
Average \# of ``think'' for   & o1-mini  & 4o  & 3.5-turbo      \\\hline
Failure case  & 15.4    & 7.4         & 9.8                    \\
Successful case & 5.3    & 2.6         & 6.1                    \\
\bottomrule[1.5pt]
\end{tabular}
}
\caption{ChatGPT ``think'' more when fail.}
\label{tab:numberThink}
\end{table}

\autoref{fig:fullLogOverthinking} shows a failure case in decision making task. The environment description is provided to the LLMs: ``You are in the middle of a room. Looking quickly around you, you see a cabinet 4, a cabinet 3, a cabinet 2, a cabinet 1, a countertop 1, a garbagecan 1, a handtowelholder 2, a handtowelholder 1, a sinkbasin 2, a sinkbasin 1, a toilet 1, a toilet paperhanger 1, and a towelholder 1''. The LLMs are asked to clean some cloth and put it in the coutertop. The correct way to do this is to first find the cloth, then go to the sinkbasin to clean it, and finally put the clean cloth on the countertop. In \textit{Initial response generation}, LLMs balance the number of ``think'' and specific action to gradually achieve the goal. ``think'' serves as a bridge between previous and next steps. Nevertheless, in \textit{Refinement}, LLMs generate much more ``think'' in order to take more caution than the first trial. Such behavior unfortunately leads to failure by looping in ``think''. 

\autoref{fig:fullLogPerfectionismBias} shows an example of perfectionism bias. The environment description is provided to the LLMs: ``You are in the middle of a room. Looking quickly around you, you see a armchair 1, a cabinet 4, a cabinet 3, a cabinet 2, a cabinet 1, a drawer 5, a drawer 4, a drawer 3, a drawer 2, a drawer 1, a dresser 1, a garbagecan 1, a safe 1, a shelf 12, a shelf 11, a shelf 10, a shelf 9, a shelf 8, a shelf 7, a shelf 6, a shelf 5, a shelf 4, a shelf 3, a shelf 2, a shelf 1, a sidetable 1, and a sofa 1.'' The LLMs are asked to find two pillows and put them in the sofa. The correct way to do this is to first find one pillow and put it in the sofa, then find another pillow and put it in the sofa. In \textit{Initial response generation}, LLM successfully completes the task by picking up two pillows one after the other. However, it wants to improve efficiency by picking up two pillows at the same time. This behavior instead leads to failure because the environment restricts it from doing so.

\autoref{fig:fullLogCognitiveOverload} shows an example of cognitive overload. The environment description is provided to the LLMs: ``You are in the middle of a room. Looking quickly around you, you see a cabinet 4, a cabinet 3, a cabinet 2, a cabinet 1, a countertop 1, a garbagecan 1, a handtowelholder 2, a handtowelholder 1, a sinkbasin 2, a sinkbasin 1, a toilet 1, a toiletpaperhanger 1, and a towelholder 1.''
 The LLMs are asked to clean some soapbar and put it in the cabinet. The correct way to do this is to first find the soapbar, then go to the sinkbasin to clean it, and finally put it on the countertop. When processing complex tasks with self-correction, the input prompts often have a very long context with feedback and history behavior. For example, the \textit{Refinement} prompt has 2000+ tokens compared to 9 tokens in Yes/No question answering (for reference, the context window of GPT-3.5-turbo is 4191). When the input prompt is too long, the model needs to parse everything in limited resources, which may lead to forgetting or overlooking some critical information stored somewhere in the long prompt. In our scenario, LLM forgets the significant syntax formulation stored somewhere in the long prompt (e.g., the correct format is ``in\textbackslash on'' rather than ``in''). This directly leads to the task failure. 
 
\clearpage
\subsubsection{Reasoning}
\label{app:reasoning}

\begin{figure*}[htbp!]
\centering
  \includegraphics[width=0.9\linewidth]{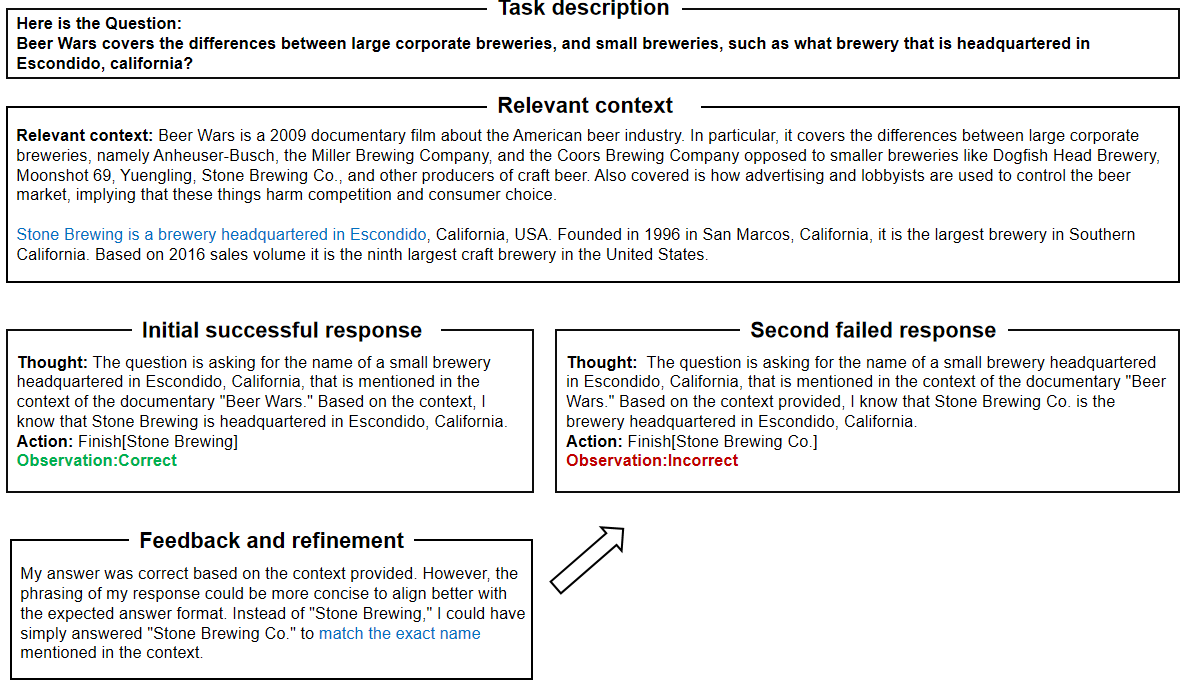}
  \caption {Full log analysis of cognitive overload in reasoning task.}
  \label{fig:fullLogCognitiveOverloadReasoning}
\end{figure*}

\begin{figure*}[htbp!]
\centering
  \includegraphics[width=0.9\linewidth]{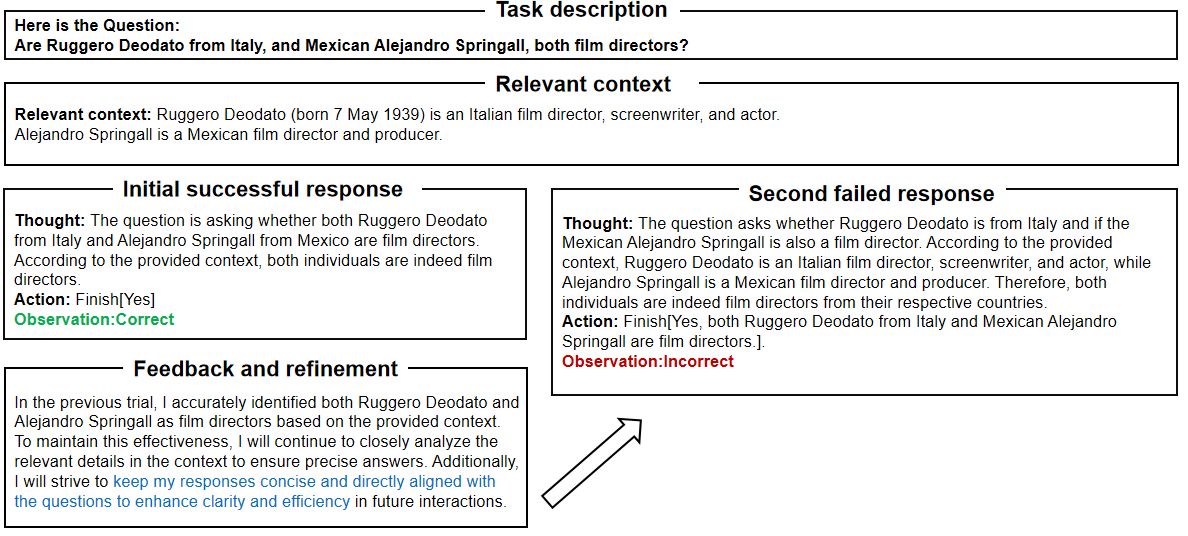}
  \caption {Full log analysis of perfectionism bias in reasoning task.}
  \label{fig:fullLogPerfectionismBiasReasoning}
\end{figure*}

In reasoning task, the LLMs are required to answer a question based on a relevant context provided. \autoref{fig:fullLogCognitiveOverloadReasoning} and \autoref{fig:fullLogPerfectionismBiasReasoning} shows the full log of cognitive overload, and perfectionism bias in reasoning task.

\autoref{fig:fullLogCognitiveOverloadReasoning} shows an example of cognitive overload. The LLMs are required to answer the question: ``What brewery is headquartered in Escondido, California?''. And a relevant context about the ``Beer Wars'' is provided. When using self-correction to handle this task, the input prompts often consist of a lengthy context, incorporating feedback and historical actions. In this example, the \textit{refinement} prompt has 1000+ tokens. LLMs overlook crucial details from the original context (e.g., the correct brewery is "Stone Brewing" rather than "Stone Brewing Co."). This omission directly leads to the task's failure. 

\autoref{fig:fullLogPerfectionismBiasReasoning} shows an example of perfectionism bias. LLM is expected to find whether Ruggero Deodato from Italy, and Mexican Alejandro Springall, both film directors. And a relevant context about Ruggero Deodato is provided. In \textit{initial response generation}, LLMs successfully complete the task by answering ``Yes''. However, LLMs want to concise the answer during self-correction, which fails because ``both Ruggero Deodato from Italy and Mexican Alejandro Springall are film directors'' is not needed. Only the answer ``Yes'' is needed.

\subsubsection{Programming}
\label{app:programming}

\begin{figure*}[htbp!]
\centering
  \includegraphics[width=\linewidth]{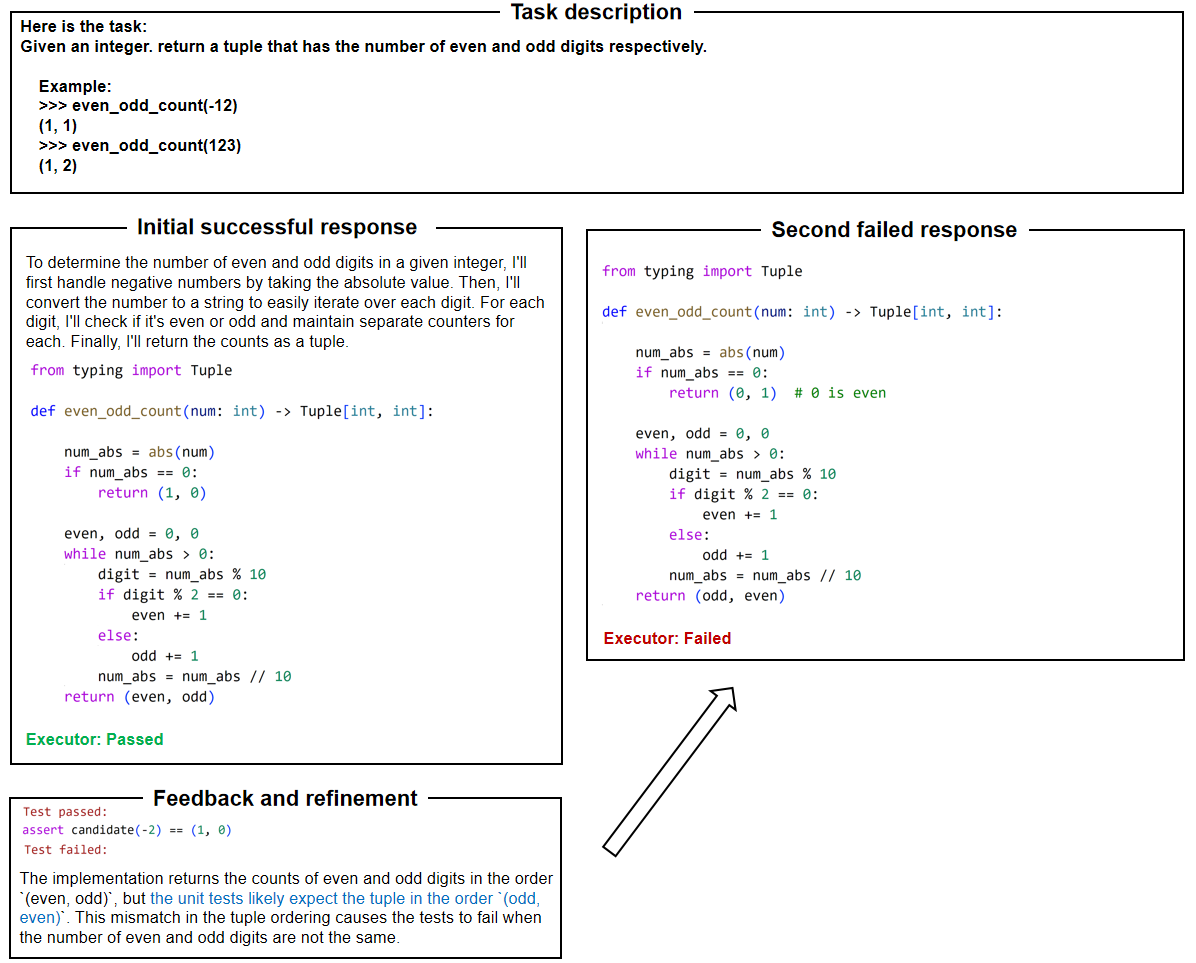}
  \caption {Full log analysis of perfectionism bias in programming task.}
  \label{fig:fullLogPerfectionismBiasProgramming}
\end{figure*}

\begin{figure*}[htbp!]
\centering
  \includegraphics[width=\linewidth]{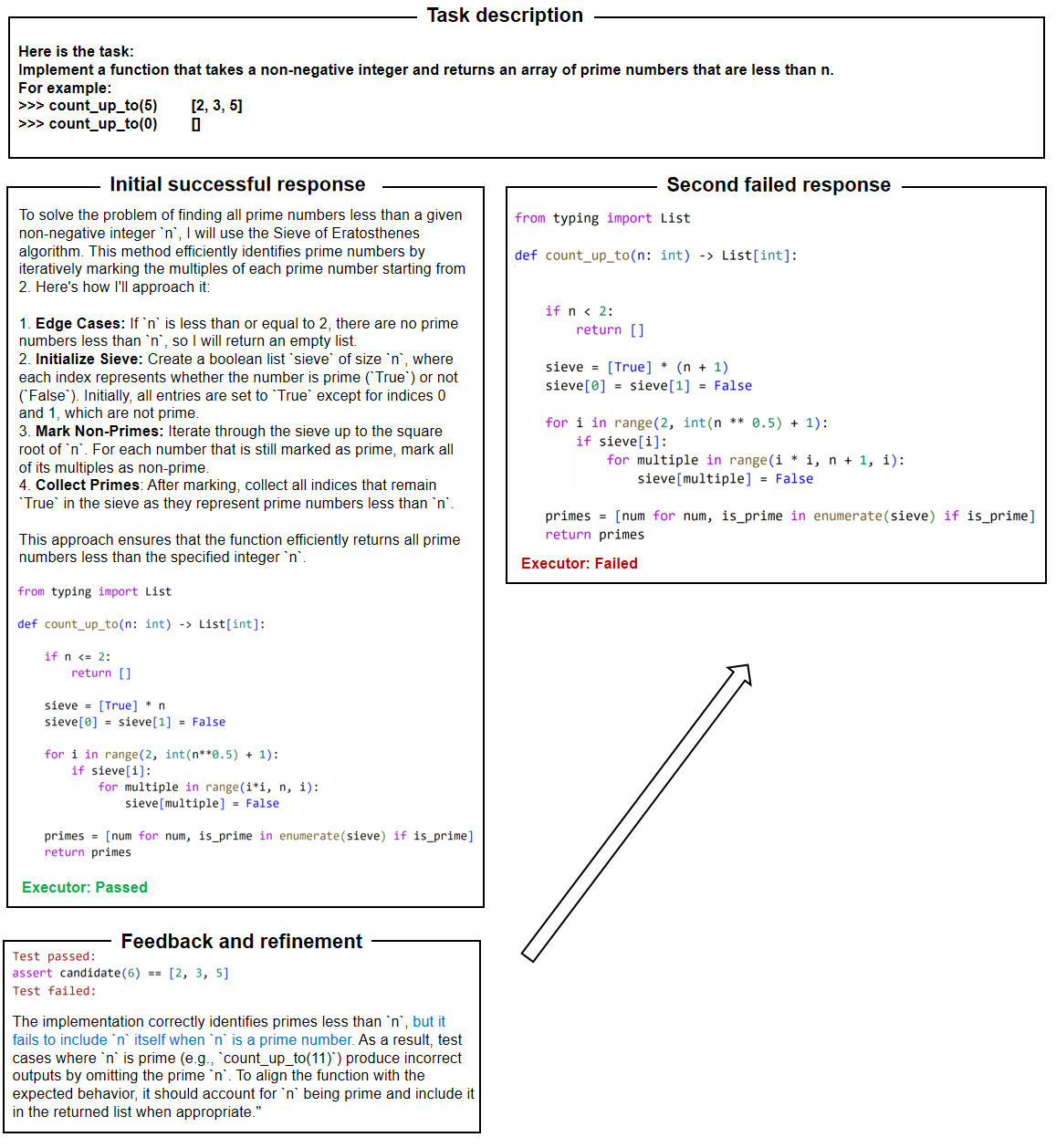}
  \caption {Full log analysis of cognitive overload in programming task.}
  \label{fig:fullLogCognitiveOverloadProgramming}
\end{figure*}

In programming task, the LLMs are required to output the code based on a description provided. And the code is tested by the executor to know whether the task is completed or not. \autoref{fig:fullLogPerfectionismBiasProgramming} and \autoref{fig:fullLogCognitiveOverloadProgramming} show the full log of cognitive overload and perfectionism bias.

\autoref{fig:fullLogPerfectionismBiasProgramming} shows an example of perfectionism bias. The LLMs are required to return a tuple that has the number of even and odd digits respectively for a given integer. In \textit{initial response generation}, LLM successfully completes the task by returning the (even, odd) tuple in correct order. However, the LLMs want to make the answers closer to the example ``assert candidate(-2) == (1, 0)''. This behavior instead leads to failure because the LLMs misunderstand ``(1, 0)'' as (odd, even) tuple. So it unnecessarily reverse the order of (even, odd) tuple, resulting in failure.

\autoref{fig:fullLogCognitiveOverloadProgramming} shows an example of cognitive overload. The LLMs are required to implement a function that takes a non-negative interger and returns an array of prime numbers that are less than an input argument n. When using self-correction to handle this task, the input prompts typically involve long contexts. For example, the \textit{refinement} prompt exceeds 1000 tokens. This can increase cognitive load of LLMs, resulting in less understanding of the original question. In this example, the LLMs forget critical details from the original question (e.g., "less than n" instead of "less than or equal to n"), leading to task failure ultimately.

\clearpage
\section{Strategies for alleviation}
\label{app:mitigation}

In this section, we add more contents to \autoref{sec:mitigation}. \autoref{app:numberSFTsamples} shows that the number of samples for SFT can be minimized. \autoref{app:internalAnswerWaveringMitigation} shows that the internal answer wavering is reduced after question repeating or SFT. \autoref{app:PACTafterPromptEngineering} and \autoref{app:PACTafterSFT} show the prompt bias is reduced after question repeating and SFT, respectively.

\subsection{The Number of Samples for SFT Can Be Minimized.}
\label{app:numberSFTsamples}

\begin{table}[tbp!]
\centering
\resizebox{0.8\linewidth}{!}{
\begin{tabular}{c|cccc}
\toprule[1.5pt]
Size    & 100 & 50  & 10  & 4   \\\hline
GPT-4o        & 0   & 0.1 & 0.5 & -   \\
GPT-3.5-turbo & 0   & 0   & 0   & -   \\
Llama-3.1-8B  & 0   & 0   & 0.1 & 0.2 \\
\bottomrule[1.5pt]
\end{tabular}
}
\caption{$\checkmark\rightarrow\crossmark(\%)$ after SFT on different size of training set. ChatGPT requires at least 10 SFT samples.}
\label{tab:numberSFTsamples}
\end{table}

We reduce the number of samples for SFT from 100 to 4 for Llama and from 100 to 10 for ChatGPT since it requires at least 10 samples\footnote{\url{platform.openai.com/docs/guides/fine-tuning}}. \autoref{tab:numberSFTsamples} shows that when we reduce the number of samples for SFT, the $\checkmark\rightarrow\crossmark(\%)$ almost unchanges and remains at minimal. This indicates that we can minimize SFT samples to reduce cost.

\begin{figure}[tbp!]
\centering
  \includegraphics[width=\linewidth]{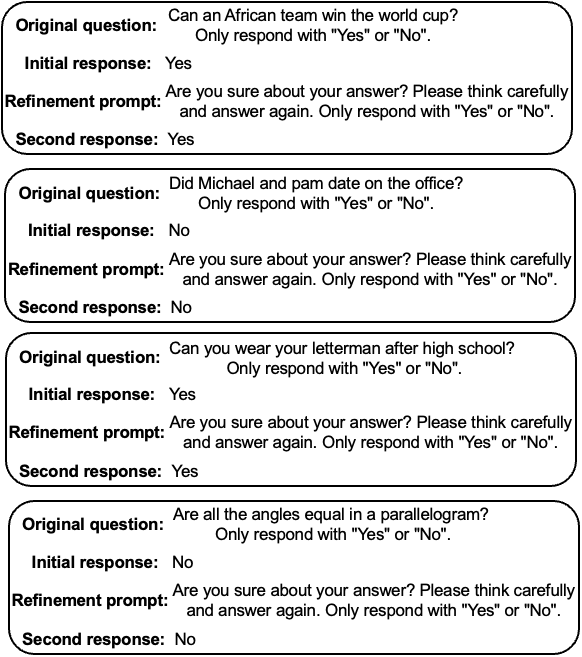}
  \caption {The 4 samples for SFT Llama.}
  \label{fig:SFTsamplesLlama}
\end{figure}

\begin{figure}[tbp!]
\centering
  \includegraphics[width=\linewidth]{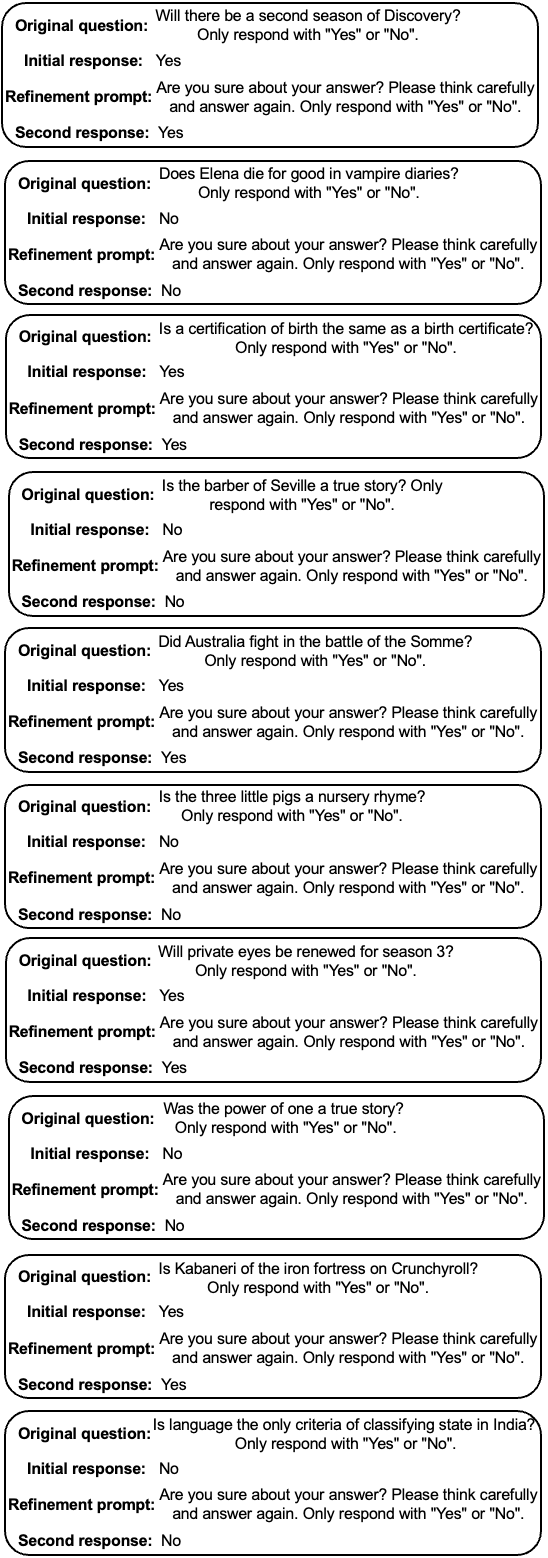}
  \caption {The 10 samples for SFT ChatGPT.}
  \label{fig:SFTsamplesChatGPT}
\end{figure}

\autoref{fig:SFTsamplesLlama} shows the 4 samples we used for SFT Llama, and \autoref{fig:SFTsamplesChatGPT} shows the 10 samples we used for SFT ChatGPT. They are all simple natural questions from BoolQ dataset \cite{clark2019boolq}.

\clearpage
\subsection{Effectiveness on other self-correction prompts}
\label{app:mitigation5prompts}

As stated in \autoref{tab:mitigate}, we conduct experiments of our strategies for all 5 self-correction prompts (described in \autoref{app:failureYesNo5prompts}). Results show that our strategies are still effective.

\autoref{apptab:acc5mitigate} and \autoref{apptab:c2i5mitigate} show that the alleviation strategies increase model's accuracy and reduce overturned correct answers. Between the two alleviation strategies, SFT further increase accuracy and reduce overturned correct answers; question repeating works in most cases while SFT works in all cases. Therefore, \textit{SFT is better than Question repeating}.

\begin{table*}[t]
\centering
\resizebox{0.9\linewidth}{!}{
\begin{tabular}{c|c|c|c|c|c}
\toprule[1.5pt]
Model         & Prompt1  & Prompt2  &Prompt3  &Prompt4  &Prompt5    \\\hline
 GPT-4o        & $79.2\hspace{0.1cm}(\downarrow 4.9)$ & $82.9\hspace{0.1cm}(\downarrow 1.2)$ & $82.3\hspace{0.1cm}(\downarrow 1.8)$& $81.9\hspace{0.1cm}(\downarrow 2.2)$& $82.3\hspace{0.1cm}(\downarrow 1.8)$\\
 + Question repeating        & $83.6\hspace{0.1cm}(\downarrow 0.5)$ & $83.6\hspace{0.1cm}(\downarrow 0.5)$ & $83.5\hspace{0.1cm}(\downarrow 0.6)$& $81.4\hspace{0.1cm}(\downarrow 2.7)$& $81.9\hspace{0.1cm}(\downarrow 2.2)$\\
 + SFT         & $87.7\hspace{0.1cm}(\uparrow 4.1)$ & $85.9\hspace{0.1cm}(\uparrow 2.3)$ & $85.9\hspace{0.1cm}(\uparrow 1.8)$& $87.1\hspace{0.1cm}(\uparrow 3.0)$& $87.1\hspace{0.1cm}(\uparrow 3.0)$\\\hline
 GPT-3.5-turbo        & $62.5\hspace{0.1cm}(\downarrow 12.1)$ & $65.5\hspace{0.1cm}(\downarrow 9.1)$ & $65.0\hspace{0.1cm}(\downarrow 9.6)$& $55.0\hspace{0.1cm}(\downarrow 24.6)$& $47.8\hspace{0.1cm}(\downarrow 26.8)$\\
 + Question repeating        & $67.4\hspace{0.1cm}(\downarrow 7.2)$ & $67.7\hspace{0.1cm}(\downarrow 6.9)$ & $63.3\hspace{0.1cm}(\downarrow 11.3)$& $67.1\hspace{0.1cm}(\downarrow 7.5)$& $53.3\hspace{0.1cm}(\downarrow 21.3)$\\
 + SFT         & $76.2\hspace{0.1cm}(\uparrow 1.6)$ & $75.3\hspace{0.1cm}(\uparrow 0.7)$ & $75.9\hspace{0.1cm}(\uparrow 1.3)$& $75.6\hspace{0.1cm}(\uparrow 1.0)$& $66.0\hspace{0.1cm}(\downarrow 8.6)$\\
\bottomrule[1.5pt]
\end{tabular}
}
\caption{$\text{ACC}_1~(\downarrow\Delta\text{ACC}) (\%)$ of alleviation strategies. Question repeating increases ACC in most cases; SFT increases ACC in all cases.}
\label{apptab:acc5mitigate}
\end{table*} 

\begin{table*}[t]
\centering
\resizebox{0.7\linewidth}{!}{
\begin{tabular}{c|c|c|c|c|c}
\toprule[1.5pt]
Model         & Prompt1  & Prompt2  &Prompt3  &Prompt4  &Prompt5    \\\hline
 GPT-4o        & $11.3$ & $5.3$ & $5.2$& $6.3$& $7.5$\\
 + Question repeating        & $6.0$ & $5.1$ & $4.8$& $6.9$& $8.0$\\
 + SFT         & $0$ & $0$ & $0$& $0.3$& $3.3$\\\hline
 GPT-3.5-turbo       & $34.0$ & $23.8$ & $26.4$& $39.3$& $61.5$\\
 + Question repeating        & $23.1$ & $25.2$ & $36.7$& $23.2$& $53.6$\\
 + SFT         & $0$ & $0$ & $0.1$& $9.3$& $17.4$\\
\bottomrule[1.5pt]
\end{tabular}
}
\caption{$\checkmark\rightarrow\crossmark (\%)$ of alleviation strategies. Question repeating reduces $\checkmark\rightarrow\crossmark (\%)$ in most cases; SFT reduces $\checkmark\rightarrow\crossmark (\%)$ in all cases.}
\label{apptab:c2i5mitigate}
\end{table*}

\subsection{Internal answer wavering is reduced by question repeating and SFT.}
\label{app:internalAnswerWaveringMitigation}

\begin{figure*}[htbp!]
    \includegraphics[width=\linewidth]{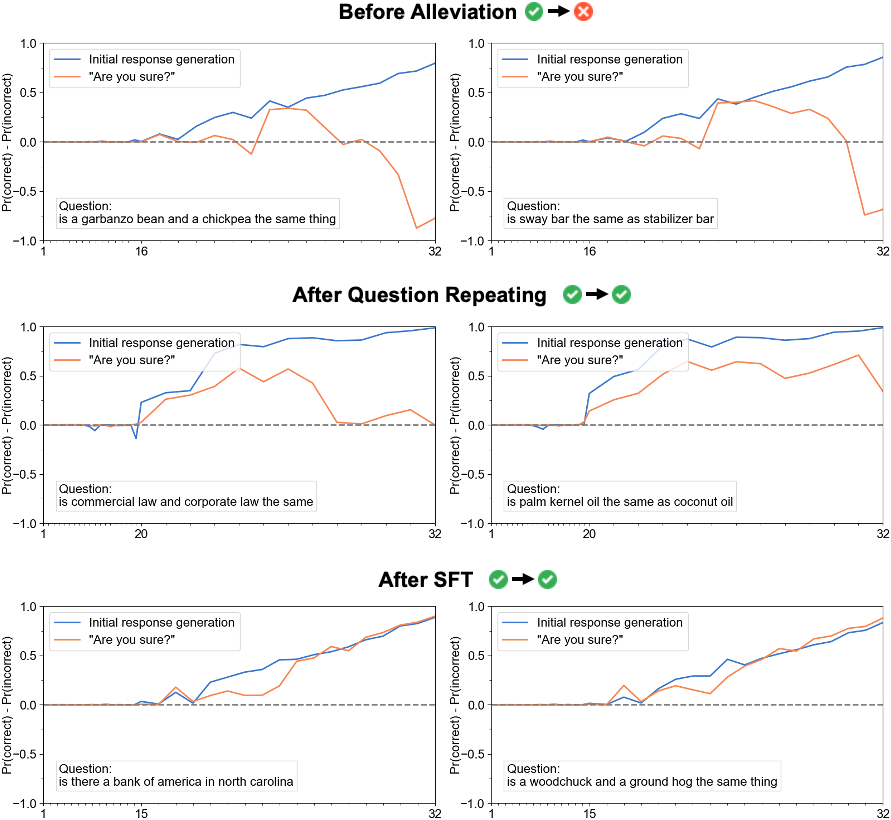}
  \caption{Internal answer wavering is reduced. Both question repeating and SFT can correct the $\checkmark\rightarrow\crossmark$ samples to $\checkmark\rightarrow\checkmark$ samples. Moreover, SFT makes the internal behaviour of \textit{Feedback and Refinement} much closer to \textit{Initial response generation} compared to question repeating, indicating that SFT is a better strategy for alleviation.}
\label{appfig:ProbingAfterMitigation}
\end{figure*}

\autoref{appfig:ProbingAfterMitigation} shows the internal answers of Llama before and after alleviation methods. The difference is on the \textit{Feedback and Refinement} curves (orange), while the \textit{Initial response generation} curves (blue) retain the same behaviour of gradually increasing with deeper layers.

Before alleviation, \textit{Feedback and Refinement} makes LLMs' internal answer waver, resulting in a wrong final answer. Specifically, the orange curve bounces up and down at 0, indicating that the model hesitates to give correct or incorrect answers.

After alleviation, both question repeating and SFT reduce the internal answer wavering during \textit{Feedback and Refinement}, and correct the $\checkmark\rightarrow\crossmark$ samples to $\checkmark\rightarrow\checkmark$ samples. Moreover, SFT makes the internal behaviour of \textit{Feedback and Refinement} much closer to \textit{Initial response generation} compared to question repeating. Specifically, the two curves in the last row of \autoref{appfig:ProbingAfterMitigation} are closer than the second row. This indicates that SFT is a better alleviation method than question repeating, which can further alleviate the negative effects of \textit{Feedback and Refinement}.

\clearpage
\subsection{Prompt bias is alleviated by question repeating}
\label{app:PACTafterPromptEngineering}

\begin{figure*}[htbp!]
    \includegraphics[width=\linewidth]{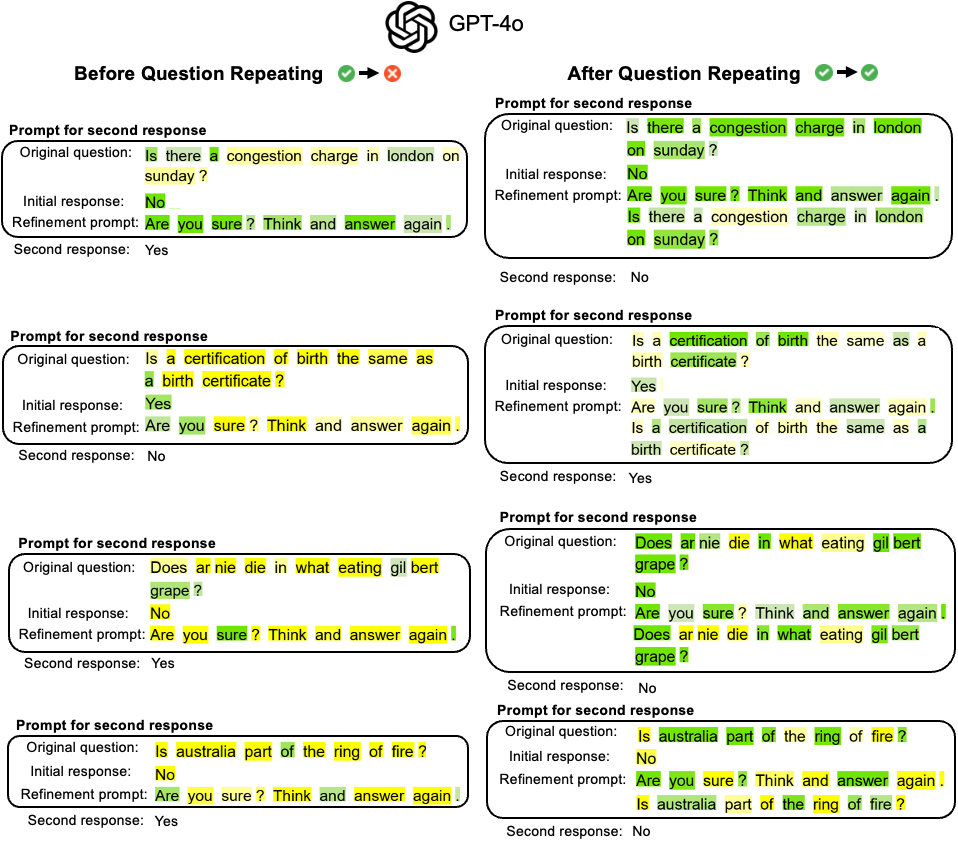}
  \caption{With question repeating, GPT-4o focus more on the original questions. \colorbox{captionup}{Greener} token means more positive contribution; \colorbox{captiondown}{Yellower} token means more negative contribution.}
\label{fig:MoreExamplesAttribAfterPromptEngineeringGpt4o}
\end{figure*}

\begin{figure*}[tbp!]
\centering
    \includegraphics[width=\linewidth]{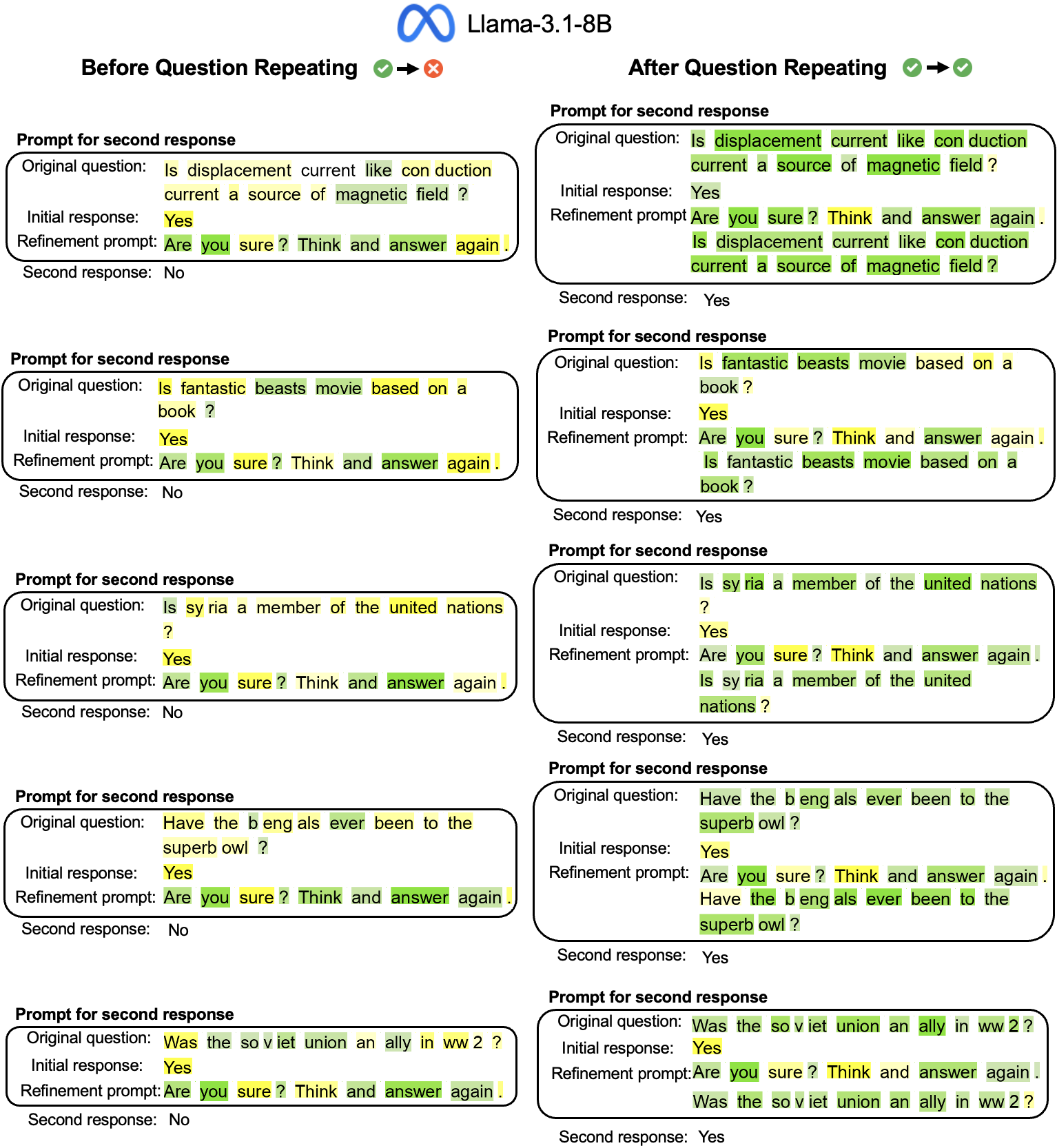}
  \caption{With question repeating, Llama-3.1-8B focus more on the original questions. \colorbox{captionup}{Greener} token means more positive contribution; \colorbox{captiondown}{Yellower} token means more negative contribution.}
\label{fig:MoreExamplesAttribAfterPromptEngineerinLlama3.1}
\end{figure*}

\autoref{fig:MoreExamplesAttribAfterPromptEngineeringGpt4o} and \autoref{fig:MoreExamplesAttribAfterPromptEngineerinLlama3.1} show more examples of input PACT before and after question repeating, for GPT-4o and Llama-3.1-8B respectively. 

For the $\checkmark\rightarrow\crossmark$ cases before question repeating, we observe that tokens in the refinement prompt are generally greener than tokens in the original question. This indicates that LLMs are biased towards refinement prompt rather than the original question itself, leading to wrong answer. After question repeating, the $\checkmark\rightarrow\crossmark$ cases become $\checkmark\rightarrow\checkmark$ cases. LLMs focus more on the original question attached to the end of the refinement prompt, which eliminates the undesirable effects of self-correction.

Specifically, for the first example in \autoref{fig:MoreExamplesAttribAfterPromptEngineeringGpt4o}, GPT-4o overturns the initial correct answer of the question ``Is there a congestion charge in london on sunday?'' before question repeating. We observe that the refinement prompt ``Are you sure? Think and answer again'' is generally greener than the original question. This means GPT-4o focuses more on the distracting refinement prompt rather than the original question when answering. After question repeating, the original question, either in the beginning of the prompt for second response or in the refinement prompt, becomes greener. This means GPT-4o correctly focuses on the original question when answering. The second response is therefore correct.

\clearpage
\subsection{Prompt bias is alleviated by SFT}
\label{app:PACTafterSFT}

\begin{figure*}[tbp!]
\centering
    \includegraphics[width=\linewidth]{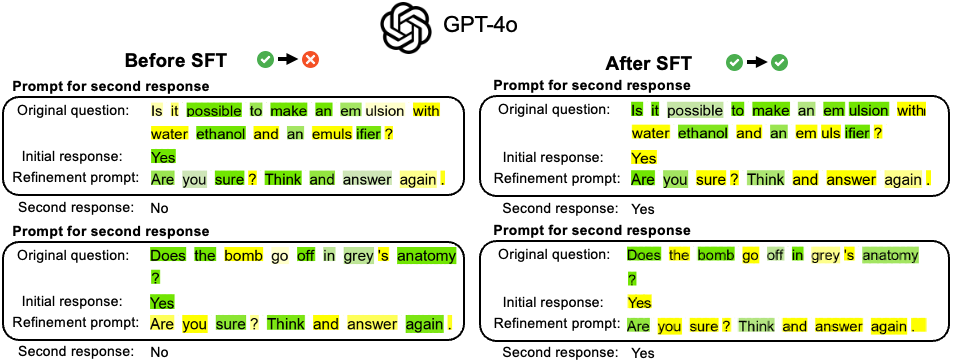}
  \caption{With SFT, GPT-4o focus more on the original questions. \colorbox{captionup}{Greener} token means more positive contribution; \colorbox{captiondown}{Yellower} token means more negative contribution.}
\label{fig:MoreExamplesAttribAfterSFTGpt4o}
\end{figure*}

\begin{figure*}[htbp!]
\centering
    \includegraphics[width=\linewidth]{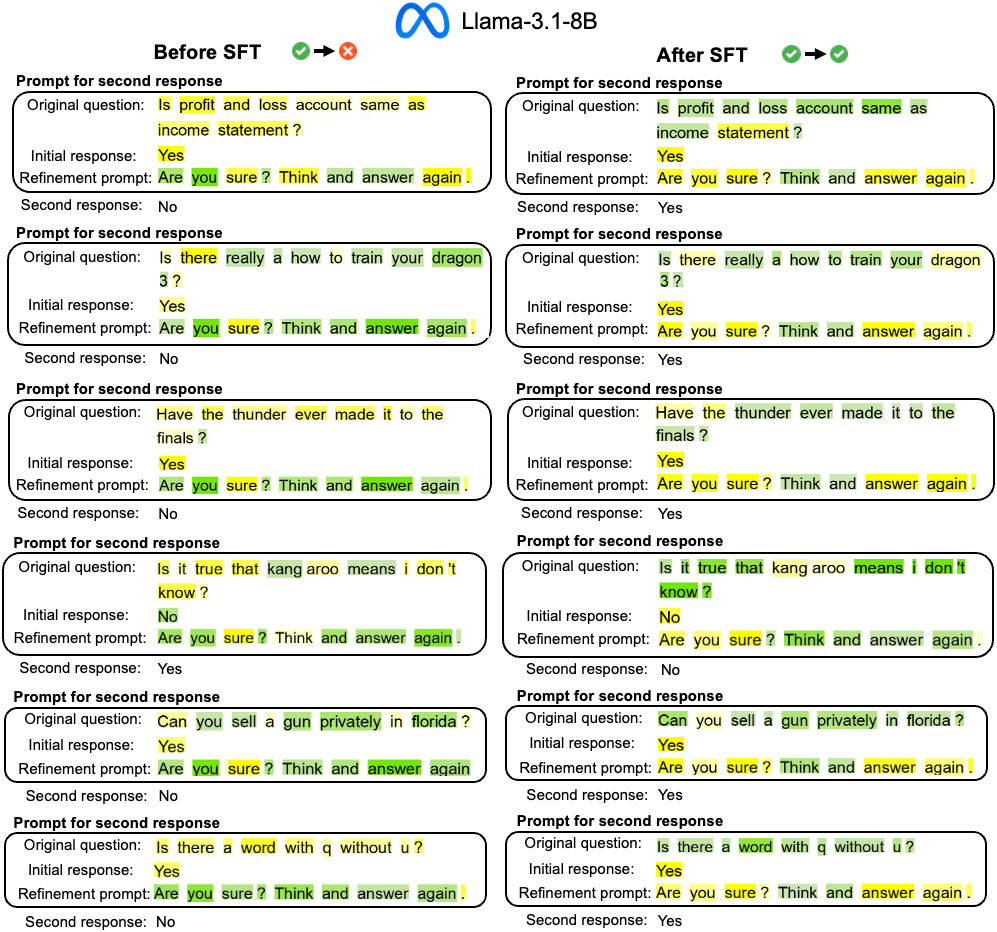}
  \caption{With SFT, Llama-3.1-8B focus more on the original questions. \colorbox{captionup}{Greener} token means more positive contribution; \colorbox{captiondown}{Yellower} token means more negative contribution.}
\label{fig:MoreExamplesAttribAfterSFTLlama3.1}
\end{figure*}

\autoref{fig:MoreExamplesAttribAfterSFTGpt4o} and \autoref{fig:MoreExamplesAttribAfterSFTLlama3.1} show more examples of input PACT before and after SFT, for GPT-4o and Llama-3.1-8B respectively. Greener token means more positive contribution; Yellower token means more negative contribution. Same as question repeating, the $\checkmark\rightarrow\crossmark$ cases become $\checkmark\rightarrow\checkmark$ cases after SFT. LLMs focus more on the original question rather than refinement prompt. This behavior corrects the prompt bias which can lead to wrong answer.

Specifically, for the first example in \autoref{fig:MoreExamplesAttribAfterSFTGpt4o}, GPT-4o overturns the initial correct answer of the question ``Is it possible to make an emulsion with water ethanol and an emulsifier?'' before SFT. We observe that the refinement prompt ``Are you sure? Think and answer again'' is generally greener than the original question. Indeed, the original question contains 8 yellow tokens while the refinement prompt contains 3 yellow tokens. This means GPT-4o focuses more on the distracting refinement prompt rather than the original question when answering. After SFT, the original question becomes greener (i.e., green tokens increase from 8 to 11) and the refinement prompt becomes yellower (i.e., yello tokens increase from 3 to 6). This means GPT-4o focuses more on the original question when answering, which is sufficient to give the second correct answer.

Similar for Llama-3.1-8B, for the first example in \autoref{fig:MoreExamplesAttribAfterSFTLlama3.1}, Llama-3.1-8B overturns the initial correct answer of the question ``Is profit and loss account same as income statement?'' before SFT. We observe that the refinement prompt ``Are you sure? Think and answer again'' is generally greener than the original question. Indeed, all tokens in the original question are yellow while the refinement prompt contains 5 green tokens. This means Llama-3.1-8B focuses more on the distracting refinement prompt rather than the original question when answering. After SFT, the original question becomes greener (i.e., green tokens increase from 0 to 9) and the refinement prompt becomes yellower (i.e., yello tokens increase from 4 to 7). This means Llama-3.1-8B focuses more on the original question when answering, which is sufficient to give the second correct answer.

\end{document}

%% file: preamble.tex
\usepackage{CJKutf8}
\usepackage{CJK}
\usepackage{verbatim}
\usepackage{multicol}
\usepackage{multirow}
\usepackage{xspace}
\usepackage{dsfont}
\usepackage{tcolorbox}
\usepackage{amsmath}
\usepackage{amssymb}

\usepackage{booktabs}
\usepackage{threeparttable}
\usepackage{tabularx}
\usepackage{stfloats}

\usepackage{soul}
\usepackage{xcolor}
\usepackage{fdsymbol}

\usepackage{algorithm}
\usepackage{algorithmic}

\newcommand{\ctext}[2]{\colorlet{cl}{#1}\sethlcolor{cl}\hl{#2}}

\definecolor{mygreen}{RGB}{11,141,10}
\definecolor{myred}{RGB}{223,68,52}
\definecolor{myblue}{RGB}{70,130,180}
\definecolor{mydeepblue}{RGB}{65,105,225}
\definecolor{myviolet}{RGB}{97,0,138}
\definecolor{myburgundy}{RGB}{110,10,30}
\definecolor{myblue2}{RGB}{0,105,148}
\definecolor{iceblue}{RGB}{173, 216, 230}
\definecolor{puregreen}{RGB}{0, 70, 0}

\definecolor{grayhighlight}{RGB}{250,250,227}

\definecolor{target}{HTML}{F47983}
\definecolor{control}{HTML}{3E87CD}
\definecolor{credibility}{HTML}{B98AC9}
\definecolor{logical}{HTML}{93C572}
\definecolor{emotional}{HTML}{F9EAC3}

\newcommand{\placeholder}[1]{\ctext{yellow!40}{#1}}

\newenvironment{packeditemize}{
\begin{list}{$\bullet$}{
\setlength{\labelwidth}{8pt}
\setlength{\itemsep}{0pt}
\setlength{\leftmargin}{\labelwidth}
\addtolength{\leftmargin}{\labelsep}
\setlength{\parindent}{0pt}
\setlength{\listparindent}{\parindent}
\setlength{\parsep}{0pt}
\setlength{\topsep}{3pt}}}{\end{list}}

\usepackage{pifont}
\newcommand{\crossmark}{\text{\ding{55}}}
\renewcommand{\checkmark}{\text{\ding{51}}}